\def\arxivpreprint{}
\documentclass{article} 
\usepackage{iclr2027_conference,times}

\usepackage{amsmath,amsfonts,bm}

\def\eqref#1{equation~\ref{#1}}

\def\1{\bm{1}}

\DeclareMathAlphabet{\mathsfit}{\encodingdefault}{\sfdefault}{m}{sl}
\SetMathAlphabet{\mathsfit}{bold}{\encodingdefault}{\sfdefault}{bx}{n}

\usepackage{hyperref}
\usepackage{url}
\usepackage{amssymb}
\usepackage{graphicx}
\usepackage{booktabs}
\usepackage{array}
\usepackage{multirow}
\usepackage{longtable}
\usepackage{xcolor}
\definecolor{MyDarkRed}{rgb}{0.8,0.02,0.02}
\definecolor{royalpurple}{rgb}{0.47, 0.32, 0.66}
\colorlet{mylinkcolor}{royalpurple} 
\colorlet{mycitecolor}{royalpurple}
\colorlet{myurlcolor}{MyDarkRed}
\hypersetup{
  citecolor  = mycitecolor,
  linkcolor = mylinkcolor,
  urlcolor = myurlcolor,
  colorlinks = true
}
\usepackage{colortbl}
\usepackage{siunitx}
\definecolor{hmg}{HTML}{1896A6}
\definecolor{grp}{HTML}{F2F4F6}
\definecolor{upc}{HTML}{1896A6}
\definecolor{dnc}{HTML}{CE595C}
\usepackage[skins]{tcolorbox}
\usepackage{alltt}
\definecolor{casebrown}{HTML}{D98A8D}
\definecolor{caseback}{HTML}{FDF5F5}
\definecolor{caseteal}{HTML}{1896A6}
\definecolor{casebackg}{HTML}{EEF7F8}
\definecolor{failred}{HTML}{CE595C}
\definecolor{okgreen}{HTML}{1896A6}
\definecolor{failtext}{HTML}{B0413F}
\definecolor{oktext}{HTML}{117480}
\definecolor{basetitle}{HTML}{B85C60}
\newtcolorbox{casebox}[1]{enhanced, colback=caseback, colframe=casebrown,
  colbacktitle=basetitle, coltitle=white, arc=2mm, boxrule=0.7pt, left=4pt,
  right=4pt, top=3pt, bottom=3pt, toptitle=2pt, bottomtitle=2pt,
  fonttitle=\bfseries\normalsize, title={#1}}
\newtcolorbox{ourbox}[1]{enhanced, colback=casebackg, colframe=caseteal,
  colbacktitle=oktext, coltitle=white, arc=2mm, boxrule=0.7pt, left=4pt,
  right=4pt, top=3pt, bottom=3pt, toptitle=2pt, bottomtitle=2pt,
  fonttitle=\bfseries\normalsize, title={#1}}
\newlength{\cfw}
\newcommand{\cframe}[3][white]{\begin{minipage}[t]{\cfw}\centering
  {\scriptsize #2}\\[1pt]{\setlength{\fboxsep}{0pt}\setlength{\fboxrule}{1.2pt}%
  \fcolorbox{#1}{white}{\includegraphics[width=\dimexpr\cfw-2.4pt\relax]{#3}}}%
  \end{minipage}}
\newcommand{\fail}[1]{\textcolor{failtext}{#1}}
\newcommand{\good}[1]{\textcolor{oktext}{#1}}
\usepackage{wrapfig}
\usepackage{float}
\usepackage{subcaption} 

\newcommand{\code}[1]{\texttt{#1}}
\usepackage{xspace}
\newcommand{\sys}{\texttt{\textbf{DynaHarness}}\xspace}
\newcommand{\pizero}{$\pi_{0.5}$}
\newcommand{\refuse}{\bot}

\title{DynaHarness: A Dynamic Physical\\ Harness for Self-Evolving Robot Agents}

\author{%
\textbf{Haoyuan Deng}$^{1}$
\quad
\textbf{Jiebin Liu}$^{1}$
\quad
\textbf{Tengxiao Zhang}$^{1}$
\quad
\textbf{Langning Yan}$^{1}$
\\
\textbf{Hongye Cao}$^{2}$
\quad
\textbf{Ziwei Wang}$^{1\dagger}$
\\[2mm]
$^{1}$Nanyang Technological University
\quad
$^{2}$Nanjing University
\\
\texttt{\{haoyuan.deng, ziwei.wang\}@ntu.edu.sg}
\\
$^{\dagger}$Corresponding author
}

\ifdefined\arxivpreprint
  \iclrfinalcopy
\fi
\begin{document}

\maketitle
\ifdefined\arxivpreprint
  \lhead{Preprint}
\fi

\begin{abstract}
Pretrained robot policies provide useful action priors, but long-horizon manipulation still requires coordination between semantic reasoning and physical execution. 
Semantic reasoning operates at a coarser timescale than physical interaction, while episode-level failures provide limited guidance on which system component should be revised. 
We propose \sys{}, a dynamic physical harness that couples semantic reasoning with physical governance through a shared execution contract and turns failure evidence into validated capability revisions. 
To be more specific, the slow brain proposes capabilities and symbolic arguments, while the fast brain grounds and monitors commands, refuses unresolved actions, substitutes capabilities, and requests replans when needed. 
The physical execution contract bounds each accepted command and records execution evidence across analytic skills, recovery skills, and the frozen VLA. 
Failure attribution localizes faults in these records and directs targeted revisions of reusable capabilities or execution mechanisms. Paired regression checks govern admission or rejection, closing the self-evolution loop. 
On LIBERO-Pro, \sys{} achieves $75.2\%$ on 800 newly sampled initial states, compared with $17.5\%$ for the frozen policy. 
With the same capability library, full dynamic execution reaches $74.0\%$ versus $63.9\%$ under nominal one-step replanning.
This demonstrates the value of \sys{} as a dynamic physical harness that governs how existing capabilities are grounded, monitored, and coordinated during execution. Our project page is at \url{https://denghaoyuan123.github.io/Dynaharness_page/}.
\end{abstract}

\edef\mainSavedParskip{\the\parskip}
\edef\mainSavedTextfloatsep{\the\textfloatsep}
\edef\mainSavedFloatsep{\the\floatsep}
\edef\mainSavedIntextsep{\the\intextsep}
\setlength{\parskip}{3pt}
\setlength{\textfloatsep}{12pt plus 2pt minus 2pt}
\setlength{\floatsep}{8pt plus 2pt minus 2pt}
\setlength{\intextsep}{10pt plus 2pt minus 2pt}
\raggedbottom

\section{Introduction}

Vision-language-action (VLA) models transfer visual and action priors to diverse
manipulation tasks~\citep{rt2,openvla,pi0,pi05}. A robot agent also needs executable
capabilities, progress monitoring, verification, recovery and a way to improve
after failure. Agentic harnesses extend pretrained policies with high-level
reasoning and tools without retraining~\citep{harnessvla2026,embodiedskills2026,eta2026,
showharness2026,phyagentos}.
Yet semantic reasoning operates at a coarser timescale than physical interaction,
and episode-level outcomes alone offer little guidance on which component to revise.

Existing embodied harnesses address these problems in two directions.
In \emph{execution orchestration}, a high-level agent decomposes tasks, selects
analytic tools or learned policies, and uses verification or recovery
~\citep{harnessvla2026,embodiedskills2026,eta2026,showharness2026,
harnesswam2026,phyagentos}.
This improves compositionality and policy reuse, but contact, progress and
intermediate success can arise between the agent's coarse planning decisions.
An explicit authority must therefore govern the running command.
In \emph{self-improvement}, failures guide program repair, skill acquisition,
recovery construction or capability-library expansion
~\citep{asd2024,aspire2026,gap2026,insight2026,enpire2026,zetta2026}.
These approaches improve a fixed library, yet a task-level failure may arise
in planning, grounding, capability execution, verification or recovery.
Localization directs a targeted revision; regression checks determine whether
it meets the system's admission criteria.
\begin{figure}[t]
\centering
\includegraphics[width=0.99\textwidth]{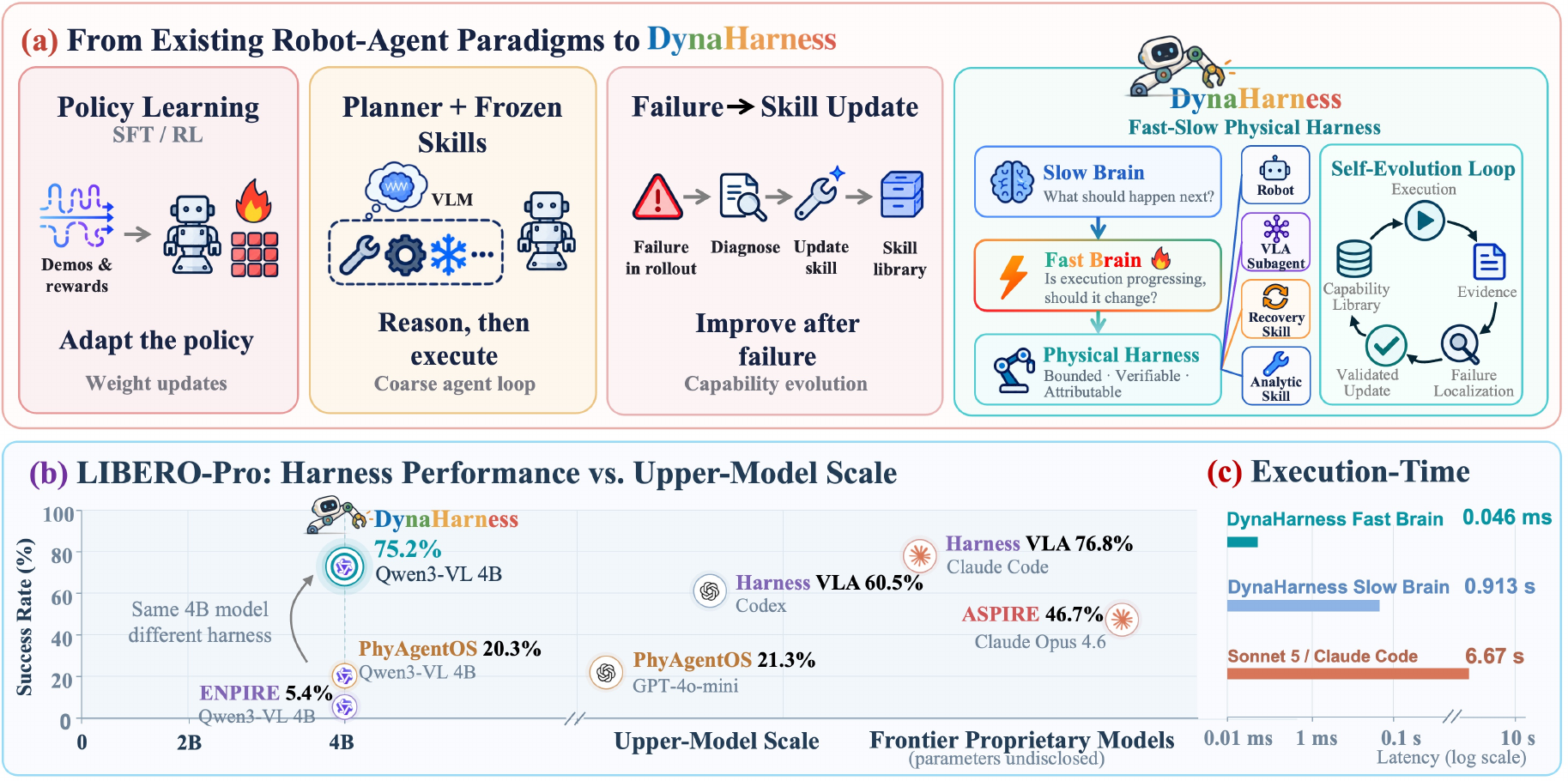}
\caption{\textbf{\sys{} at a glance.} (a) Robot-agent paradigms and \sys{}. (b)
LIBERO-Pro success against upper-model scale (Table~\ref{tab:main}). (c) Median
scheduler and Qwen call times (Table~\ref{tab:latency}); Sonnet/Claude Code
slow-brain pilot (Table~\ref{tab:upper}).}
\label{fig:teaser}
\end{figure}

The design question is how to govern physical execution and turn its evidence
into validated revisions of reusable capabilities. Prior work offers fast
progress estimation, intervention and hybrid skills~\citep{
harnesswam2026,uniintervene2026,harnessvla2026,embodiedskills2026};
we connect their execution evidence to capability revision through one command record.

We present \sys{} (Fig.~\ref{fig:teaser}), a dynamic physical harness that
connects semantic reasoning and physical governance through an execution contract.
The Qwen3-VL-4B slow brain proposes a capability and symbolic arguments.
At $2$\,Hz, the fast brain grounds and monitors commands, refuses unresolved
actions, substitutes capabilities and requests replans as the $20$\,Hz controller
executes. The contract bounds commands by budget and lease and records decisions
and outcomes across analytic skills, recovery skills and the frozen VLA.
Offline, failure attribution uses these records to locate faults within $N$
ordered layers. The resulting evidence directs targeted revisions of reusable
capabilities or execution mechanisms; paired regression checks govern admission.
The record also supplies decision data for learning the fast brain.
On LIBERO-Pro~\citep{liberopro2025}, \sys{} reaches $74.25\%$ on the development
block, compared with $20.3\%$ for PhyAgentOS with the same frozen policy and
Qwen3-VL-4B. After freezing, we generate $800$ new states of the same cells:
\sys{} succeeds on $75.2\%$, versus $17.5\%$ for the policy. Four frozen
snapshots retain their development ranking on these states, gaining $14.4$
percentage points against $14.3$ during development. A separate paired ablation
reduces success from $74.0\%$ to $16.6\%$ when seven analytic contact skills
are removed. With the same skills, nominal one-step replanning scores $63.9\%$
and frozen sequencing $63.8\%$, compared with $74.0\%$ for Full.

The main contributions of this paper are summarized as follows:
\begin{itemize}
    \item \textbf{Structured physical capabilities.}
    A shared physical execution contract grounds, bounds and refuses commands
    across analytic skills, recovery skills and a frozen VLA, recording evidence
    for revision.

    \item \textbf{Fast-slow capability execution.}
    On-demand semantic reasoning and fast physical governance coordinate
    capability execution through monitoring, substitution, recovery and replanning
    beyond nominal one-step replanning.

    \item \textbf{Attribution-driven self-evolution.}
    Execution records localize failures and direct targeted revisions of reusable
    capabilities or execution mechanisms. Paired regression checks govern their
    admission; frozen snapshots retain development gains on new initial states.
\end{itemize}

\section{Related Work}

\paragraph{Robot agents over pretrained policies}
Code as Policies turns language into executable robot programs~\citep{codeaspolicies2023};
VoxPoser and ReKep ground language in 3D value maps and relational keypoint
constraints, respectively~\citep{voxposer2023,rekep2025}.
Harness VLA composes analytic primitives with a frozen VLA for non-contact
motion and contact-rich execution~\citep{harnessvla2026}. EmbodiedSkills
organizes observation, planning, preflight, execution, verification and
recovery~\citep{embodiedskills2026}; ETA and Show-Harness expose executable tools
or semantic actions to a high-level agent~\citep{eta2026,showharness2026}.
Thea uses execution outcomes for retry and termination~\citep{thea2026}, while
PhyAgentOS provides runtime scheduling, verification and safety services
separately from cognitive planning~\citep{phyagentos}.
HarnessWAM combines high-frequency progress estimation with slower task
management~\citep{harnesswam2026}, and UniIntervene separates intervention
triggering from recovery~\citep{uniintervene2026}. Related work learns an outer
loop over a frozen policy~\citep{robustexec2026}, studies controlled
promotion~\citep{agenticloop2026,harbor2026}, and develops non-oracle perception
or physical memory~\citep{real2026,roboexp2024}.
\sys{} builds on these decompositions with a command contract that records
grounding, refusal, execution bounds and completion for later improvement.

\paragraph{Self-improving robot agents}
Agentic Skill Discovery proposes tasks and learns corresponding
skills~\citep{asd2024}. Neuro-symbolic recovery localizes plan execution errors
and replans after failure~\citep{kalithasan2024}. ASPIRE repairs execution traces and searches over
programs~\citep{aspire2026}, while GaP represents a policy as an editable graph
and rehearses changes before deployment~\citep{gap2026}. InSight acquires missing
primitives and evaluates composition and retention~\citep{insight2026}; ENPIRE
studies policy improvement on physical hardware~\citep{enpire2026}.
Zetta combines high-frequency runtime critics, recovery, hierarchical failure
diagnosis and validation-gated updates~\citep{zetta2026}. \sys{} uses command-level
provenance, refusal and termination records for automated diagnosis and admission
of changes.

\section{Method}
\label{sec:method}
\setlength{\abovedisplayskip}{4pt}\setlength{\belowdisplayskip}{4pt}
\setlength{\abovedisplayshortskip}{2pt}\setlength{\belowdisplayshortskip}{2pt}

\begin{figure}[t]
\centering
\includegraphics[width=0.98\textwidth]{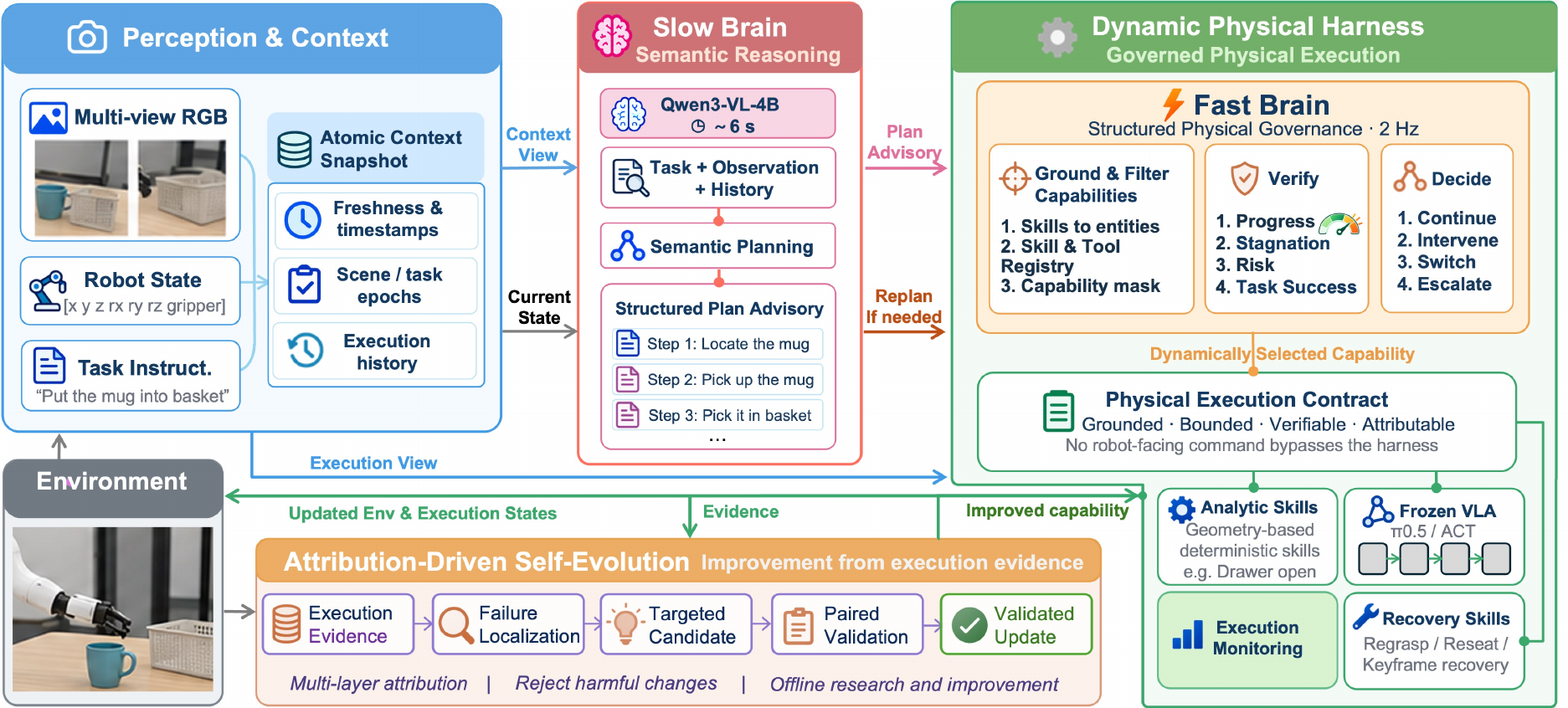}
\caption{\textbf{Overview of \sys{}.} The slow brain plans; the fast brain governs
execution; the offline loop evaluates updates from recorded evidence.}
\label{fig:runtime}
\end{figure}

\subsection{Overview}

\sys{} operates at different time scales (Fig.~\ref{fig:runtime}). The
\emph{slow brain} advises the next step. The \emph{dynamic physical harness}
grounds commands, checks admissibility, enforces budgets and leases, routes
capabilities, monitors execution, coordinates recovery and records evidence. Its
\emph{fast brain} governs the running command; its \emph{execution contract}
admits every robot-facing command. Offline, \emph{self-evolution} localizes failures
from this record and admits changes through paired validation. The parts exchange
\emph{capabilities}: bounded physical commands from a library $\mathcal{L}$ of
analytic skills, recovery skills and \code{vla\_act}, which calls a frozen VLA.
Control steps $t$ run at $20$\,Hz, fast-brain decisions $k$ at $2$\,Hz, and plan
steps $j$ on demand; $o_t = (I_t, q_t, f_t)$ contains two camera views, the
end-effector pose, gripper state and finger contact.

\subsection{Slow Brain: Semantic Reasoning}
\label{sec:slow}

The slow brain, Qwen3-VL-4B called on demand, reads an atomic context snapshot:
instruction $\ell$, scene description $S$, current observation and history $h_j$
with timestamps and scene and task epochs. It returns a structured advisory:
\begin{equation}
c_j = \Phi(\ell,\; S,\; o_t,\; h_j), \qquad
z_j = \Pi(c_j) = (m_j,\; \alpha_j), \quad m_j \in \mathcal{L}.
\label{eq:ctx}
\end{equation}
The advisory names a capability $m_j$ and symbolic arguments $\alpha_j$, but no
poses; the fast brain resolves geometry, success and refusal, and requests replans.

\begin{figure}[t]
\centering
\begin{ourbox}{\footnotesize The fast brain within one episode: \emph{push the plate
to the front of the stove}, \code{goal\_swap[5]}, seed 22}
\begin{minipage}[c]{0.6\linewidth}
\setlength{\cfw}{0.325\linewidth}
\noindent
\cframe[okgreen]{refused, step 10}{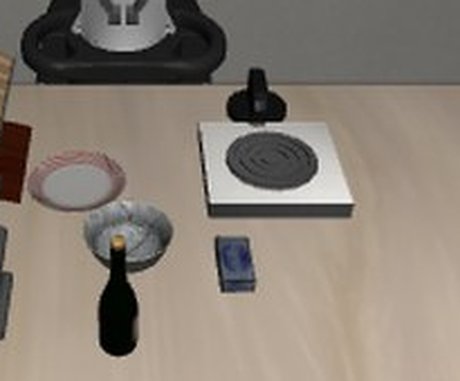}\hfill
\cframe{carried, step 144}{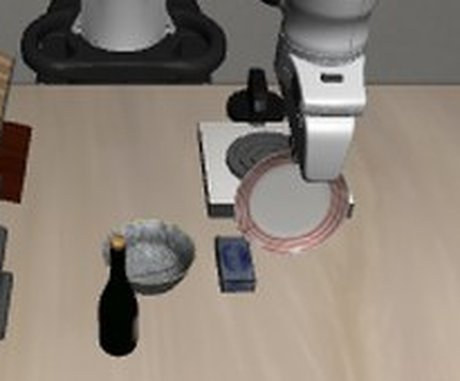}\hfill
\cframe[okgreen]{completed, step 245}{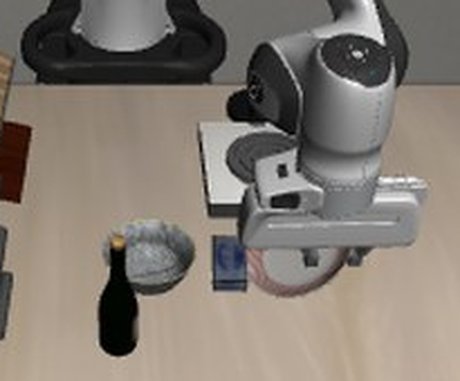}\\[1pt]
{\scriptsize\bfseries
\makebox[\cfw]{Refuse, substitute}\hfill
\makebox[\cfw]{Execute: carry}\hfill
\makebox[\cfw]{Verdict: latched}}
\end{minipage}\hfill
\begin{minipage}[c]{0.37\linewidth}\scriptsize
\textbf{Refuse, substitute.} \code{push\_object} is refused at step $10$, and a
pick and a place are substituted.\\[2pt]
\textbf{Execute.} Analytic stages grasp and carry the plate.\\[2pt]
\textbf{Complete.} The latched completion verdict ends the command at step $245$.\\[2pt]
\textbf{Success in 246 of 300 steps}; the frozen \pizero{} fails on the
same seed.
\end{minipage}
\end{ourbox}
\vspace{-3mm}
\caption{\textbf{The fast brain within one episode}; the full episode is in
Fig.~\ref{fig:case_ours_push}.}
\label{fig:example}
\end{figure}

\subsection{Fast Brain: Physical Governance}
\label{sec:fast}
\label{sec:rates}

The fast brain holds execution authority, runs deterministically at $2$\,Hz
and governs rather than replans (Fig.~\ref{fig:example}).
\textbf{Ground and filter.} An advised step becomes a command through a partial
map
\begin{equation}
\gamma_k = G(z_j,\; c_j) \;\in\; \Gamma \cup \{\refuse\}, 
\qquad \gamma = (m,\; \theta,\; B,\; \tau),
\label{eq:ground}
\end{equation}
where $\theta$ are the physical parameters, $B$ the step budget and $\tau$ the
lease. Grounding binds executable capabilities to the state estimates available
to the execution interface, checks registry entries and preconditions, and keeps
a mask of eligible capabilities. If parameters or preconditions cannot be
resolved, it returns the refusal $\refuse$ with its reason.

\textbf{Monitor.} At each decision, the fast brain reads progress, stagnation
and risk; the execution interface accumulates completion events between decisions:
\begin{equation}
v_k = v_{k-1} \;\vee\!\bigvee_{t\in\mathcal{T}_k} b_t,
\qquad v_{-1} = \mathrm{false},
\label{eq:verify}
\end{equation}
where $\mathcal{T}_k$ indexes completion checks since the preceding decision and
$b_t$ is the Boolean verdict. In LIBERO, it comes from the benchmark success
predicate. Latching retains detected events until the fast brain reads them;
detection depends on the interface's check frequency.

\textbf{Decide.} The fast brain continues execution, invokes recovery, switches
to another eligible capability or \code{vla\_act}, or requests a slow-brain
replan. Commands end on local completion or failure, latched verdict $v_k$,
budget exhaustion or lease expiry.

\subsection{Physical Execution Contract}
\label{sec:contract}

No robot-facing command bypasses one contract. \textbf{Grounded:} a command
enters the physical world only if Eq.~\ref{eq:ground} resolves its parameters
and preconditions through the execution interface. \textbf{Bounded:}
it acts only while its budget remains, its lease holds and a $50$\,Hz safety
check passes, and a capability unable to afford its own completion refuses
before it starts. \textbf{Verifiable:} it ends on a recorded local status,
latched task verdict, budget or lease condition.
\textbf{Attributable:} an episode evidence store logs observations, decisions,
outcomes and reasons for planning and attribution (Appendix~\ref{app:method}).
Refusals remain in the record even when no action is dispatched, allowing
diagnosis to distinguish unresolved commands from failed capability executions.

\subsection{Capability Execution}
\label{sec:dispatch}

The selected capability executes the command: $a_t = T_m(\theta, o_t)$ for an
analytic skill, geometric and deterministic, or a recovery skill such as a
regrasp, a re-seat or a keyframe recovery, and $a_t = \pi(o_t, \ell)$ for
\code{vla\_act} with the frozen VLA $\pi$ (\pizero{} in all experiments). All
spend the same budget under the same lease, run under the fast brain's
monitoring and share the task-verdict interface. Analytic skills cover geometric
and contact operations and can complete entire tasks. The VLA provides a learned
route, including steps without an eligible analytic capability
(Section~\ref{sec:vlausage}).
Switching executors preserves the command interface: analytic and learned
actions remain subject to the same grounding, monitoring and termination rules.

\subsection{Failure Attribution}
\label{sec:attribution}

Offline self-evolution begins with execution evidence. For a failed episode $i$
with evidence store $E_i$, the diagnostic procedure returns the first matching
label among $N$ ordered diagnostic layers,
\begin{equation}
\lambda_i = A_N(E_i) \in \{1,\dots,N\},
\label{eq:attrib}
\end{equation}
where $N$ is a hyperparameter controlling diagnostic granularity. Checks cover
infrastructure, context, planning, grounding, dispatch, capability execution,
verification and recovery, prioritizing upstream conditions (Appendix~\ref{app:method}).
Labels are assigned automatically, and the experience
$e_i = (\xi_i, \lambda_i, y_i)$ pairs its failure signature with that layer and
the outcome.

\subsection{Self-Evolution and Paired Validation}
\label{sec:evolution}

The attributed failure signature directs fault reproduction and a targeted
revision of a reusable capability or execution mechanism. Automated paired
validation evaluates the revision (Appendix~\ref{app:method}). The revision is
stated in physical quantities, such as a hinge radius or hand span, rather than
keyed to a task. It changes the library to $\mathcal{L}' = \mathcal{L} \oplus \kappa$;
the paired gate checks the following conditions on $\mathcal{D}$:
\begin{equation}
\textstyle\sum_{d} s_d(\mathcal{L}') \ge \sum_{d} s_d(\mathcal{L}),\;\;
\sum_{d} u_d(\mathcal{L}') \le \sum_{d} u_d(\mathcal{L}),\;\;
s_d(\mathcal{L}') \ge s_d(\mathcal{L})\;\forall d \in \mathcal{D}_\pi,\;\;
x(\mathcal{L}') = 0,
\label{eq:gate}
\end{equation}
where $s_d$ counts successes in cell $d$, $u_d$ the failures attributed to a
harness layer, $\mathcal{D}_\pi$ the cells the bare policy already wins, and $x$
contaminated episodes, paired on suite, task and seed over development seeds;
admission also requires broader regression checks. A revision that fails these
checks is not admitted; use as a development baseline does not establish admission.

\section{Experiments}
\label{sec:main}

\begin{table}[t]
\centering
\begin{minipage}[t]{0.605\textwidth}
\vspace{0pt}
\captionof{table}{\textbf{Success rate (\%) on four LIBERO-Pro cells.}}
\label{tab:main}
\centering
\scriptsize
\setlength{\tabcolsep}{4.4pt}
\renewcommand{\arraystretch}{0.98}
\begin{tabular}{@{}ll*{4}{S[table-format=2.1]}S[table-format=2.2]@{}}
\toprule
Method & Upper VLM & {Goal-T} & {Goal-S} & {10-T} & {10-S} & {Avg.} \\
\midrule
\multicolumn{7}{@{}l}{\textit{Backbone policies}}\\
OpenVLA & none & 0.0 & 0.0 & 0.0 & 0.0 & 0.0 \\
\pizero{} & none & 0.0 & 38.0 & 1.0 & 8.0 & 11.8 \\
\pizero{} (our run) & none & 19.5 & 17.5 & 20.0 & 8.0 & 16.2 \\
\pizero{}-SFT & none & 45.0 & 42.0 & 49.0 & 14.0 & 37.5 \\
Fast-WAM & none & 9.0 & 6.0 & 11.0 & 0.0 & 6.5 \\
\addlinespace[3pt]
\multicolumn{7}{@{}l}{\textit{Agentic baselines}}\\
PhyAgentOS & GPT-4o-mini & 21.0 & 35.5 & 18.0 & 10.5 & 21.3 \\
PhyAgentOS & Qwen3-VL-4B & 23.0 & 33.5 & 15.5 & 9.0 & 20.3 \\
EmbodiedSkills & Qwen3-VL-4B & 11.0 & 19.5 & 14.5 & 12.5 & 14.4 \\
ENPIRE & Qwen3-VL-4B & 10.0 & 11.5 & 0.0 & 0.0 & 5.4 \\
CaP-Agent0 & not stated & 16.8 & 25.6 & 2.4 & 5.2 & 12.5 \\
RHO & Codex/Claude Code & 55.6 & 50.6 & {--} & {--} & {--} \\
SPARK & Gemini 3.1 Pro & 14.0 & 40.0 & {--} & {--} & {--} \\
Pigey & Claude Opus-4.7 & 22.0 & 44.0 & {--} & {--} & {--} \\
Pigey & Claude Haiku-4.5 & 20.0 & 38.0 & {--} & {--} & {--} \\
Pigey & Gemini 3.5 Flash & 28.0 & 48.0 & {--} & {--} & {--} \\
Pigey & GPT-5.5 & 24.0 & 44.0 & {--} & {--} & {--} \\
VLS & not stated & 33.5 & 38.0 & 25.5 & 15.5 & 28.1 \\
Harness VLA & GPT-5.5 & 75.0 & 66.0 & 52.0 & 49.0 & 60.5 \\
Harness VLA & Claude Opus-4.7 & {\underline{87.0}} & {\underline{87.0}} & {\underline{71.0}} & {\underline{62.0}} &{\textbf{76.8}} \\
Harness VLA & Qwen3-VL-4B & 14.0 & 7.0 & 3.0 & 0.0 & 6.0 \\
Zetta & GPT-5.6sol & {\textbf{92.5}} & {\textbf{89.0}} & 63.0 & 40.0 & 71.1 \\
Zetta & Qwen3-VL-4B & 0.0 & 0.0 & 0.0 & 0.0 & 0.0 \\
ASPIRE & Claude Opus 4.6 & 45.0 & 81.0 & 38.3 & 22.6 & 46.7 \\
\midrule
\textbf{\sys{}} & Qwen3-VL-4B & 75.0 & 81.0 & {\textbf{76.0}} & {\textbf{65.0}} & {\underline{74.25}} \\
\bottomrule
\end{tabular}
\par\vspace{2pt}
\parbox{\linewidth}{\scriptsize Bold/underline: largest/second; ``--'' unreported.
\sys{}: archived aggregate $74.25\%$ on development seeds.
Protocols: Appendix~\ref{app:protocols}.}
\end{minipage}\hfill
\begin{minipage}[t]{0.37\textwidth}
\vspace{0pt}
\includegraphics[width=\linewidth]{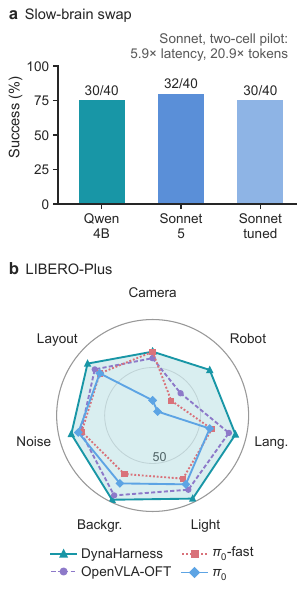}
\end{minipage}
\par\vspace{2mm}
\includegraphics[width=\textwidth]{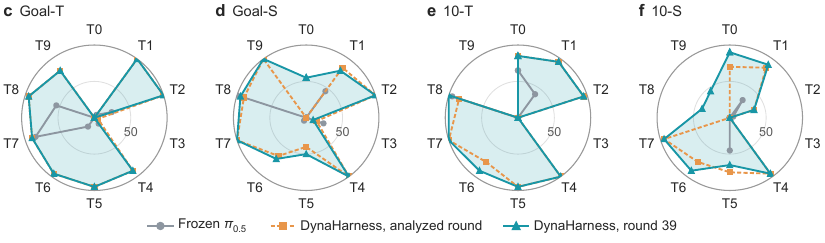}
\vspace{-6mm}
\captionof{figure}{\textbf{Success across models, tasks and perturbations.}
\textbf{a}~Sonnet~5 replacing Qwen3-VL-4B before and after prompt tuning: $40$
episodes from four cells held out from tuning; cost ratios from a separate
two-cell pilot. \textbf{b}~Seven LIBERO-Plus perturbation categories and published
references~\citep{liberoplus2025}. \textbf{c--f}~Per-task success in four
LIBERO-Pro cells ($20$ episodes per task): frozen policy versus \sys{} at the
analyzed round and round~39.}
\label{fig:profiles}
\end{table}


Experiments address three core questions:
\textbf{(1)} Does \sys{} improve task success over frozen-policy and agentic
baselines, and do development gains transfer to new initial states
(Sections~\ref{sec:literature}--\ref{sec:postselection})?
\textbf{(2)} Which execution mechanisms and capabilities support performance
(Sections~\ref{sec:fastresults}--\ref{sec:harness} and~\ref{sec:causal})?
\textbf{(3)} How do attribution and paired validation guide development updates (Section~\ref{sec:evoresults})?

\subsection{Experimental Setup}
\label{sec:setup}

\textbf{Benchmarks.} LIBERO-Pro~\citep{liberopro2025} is our main benchmark: the
Goal and LIBERO-10 suites under a task perturbation that redirects the
instruction (T) and a swap perturbation that moves the objects (S), four cells of
$10$ tasks with $20$ seeds each ($800$ episodes). Evolution uses development
seeds $21$--$40$; the early no-evolution comparison uses seeds $1$--$20$. Arms
are paired within each block. Post-selection evaluation uses $800$ newly sampled
states across $40$ task--perturbation cells and $261$ untouched official states
across $31$ cells. LIBERO-Plus~\citep{liberoplus2025} adds $10{,}030$ tasks under
seven perturbation categories using a separate configuration.

\textbf{Implementation and baselines.} Qwen3-VL-4B-Instruct is served locally;
the public \pizero{} LIBERO checkpoint~\citep{pi05} stays frozen. Our controlled
runs share the four LIBERO-Pro cells and benchmark success metric. PhyAgentOS~\citep{phyagentos}
and Harness VLA~\citep{harnessvla2026} also share \sys{}'s Qwen3-VL-4B upper model,
frozen \pizero{}, evaluation budget and $800$-episode protocol. We additionally
run PhyAgentOS with GPT-4o-mini, and ENPIRE~\citep{enpire2026}, Zetta
and EmbodiedSkills with Qwen3-VL-4B. ASPIRE and CaP-Agent0 adaptation runs
are documented in Appendix~\ref{sec:ranbaselines}. \emph{Upper VLM} denotes the model
above the policy; success counts all episodes. Analyses requiring episode records
identify the corresponding run; the earlier \emph{analyzed round} is round~30. Appendix~\ref{app:protocols}
gives baseline sources and settings.

\begin{table}[t]
\centering
\begin{minipage}[t]{0.575\textwidth}
\vspace{0pt}
\captionof{table}{\textbf{Transfer and execution ablations.}}
\label{tab:transfer_mechanism}
\centering
\scriptsize
\setlength{\tabcolsep}{4pt}
\renewcommand{\arraystretch}{1.12}
\begin{tabular}{@{}lrr@{}}
\toprule
\multicolumn{3}{@{}l}{\textbf{A. Frozen snapshots: success (\%)}}\\
Snapshot & Development & New states \\
\midrule
stat23 & 60.00 & 60.9 \\
stat28 & 66.25 & 65.9 \\
stat39 & 72.75 & 74.2 \\
\rowcolor{hmg!10}\textbf{Final} & \textbf{74.25} & \textbf{75.2} \\
\midrule
\multicolumn{3}{@{}l}{\textbf{B. Paired execution ablations}}\\
Variant & Success (\%) & $\Delta$ (pp) \\
\midrule
Full \sys{} (A2ctrl) & 74.0 & --- \\
\rowcolor{hmg!10}Nominal replanning (A2static) & 63.9 & $-10.1$ \\
Frozen sequence (A2seq) & 63.8 & $-10.3$ \\
No analytic contact skills & 16.6 & $-57.4$ \\
No recovery/intervention & 73.1 & $-0.9$ \\
Neither group + \pizero{} & 15.8 & $-58.2$ \\
Bare \pizero{} & 16.2 & --- \\
\bottomrule
\end{tabular}
\par\vspace{2pt}
{\scriptsize A: $800$ new states sampled after freezing; rank correlation $+1.00$.
B: $800$ development episodes per arm; deltas use A2ctrl ($74.0\%$).
A2static retains nominal replanning; A2seq freezes a sequence (Section~\ref{sec:causal}).}
\end{minipage}\hfill
\begin{minipage}[t]{0.4\textwidth}
\vspace{0pt}
\centering
\includegraphics[width=\linewidth]{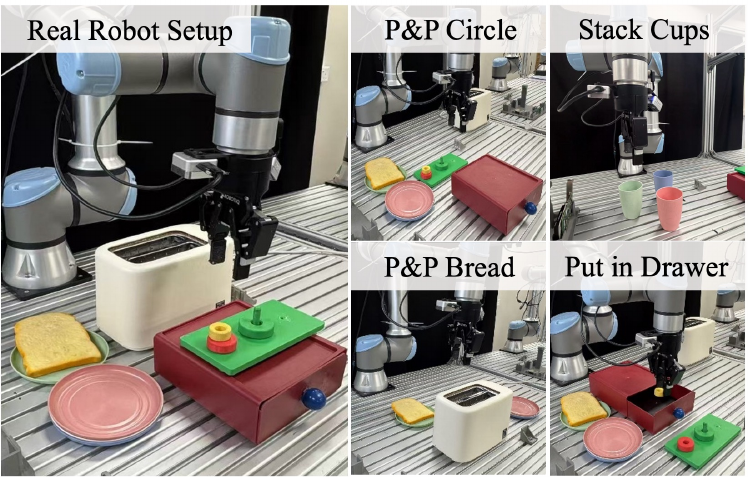}
\captionsetup{justification=raggedright,singlelinecheck=false,skip=3pt}
\captionof{figure}{\textbf{Real-world setup and tasks.} Workspace and four tasks.}
\label{fig:realrobot}
\par\vspace{2mm}
\captionof{table}{\textbf{Real-world success.}}
\label{tab:realrobot}
\scriptsize
\begin{tabular*}{\linewidth}{@{\extracolsep{\fill}}lc@{}}
\toprule
Task & Success (\%) \\
\midrule
Pick and place circle & 90.0\\
Stack cups & 80.0\\
Pick and place bread & 70.0\\
Put in the drawer & 70.0\\
\bottomrule
\end{tabular*}
\end{minipage}%
\end{table}

\subsection{Main Results}
\label{sec:literature}

\textbf{Standard LIBERO.} With settings adapted to standard LIBERO, \sys{}
scores $98.05\%$, versus $98.0\%$ for the policy reference
(Fig.~\ref{fig:libero}). Both are near ceiling on these tasks; perturbed
LIBERO-Pro therefore provides the main execution comparison. Its development
and post-selection evaluations use the final configuration
(Appendix~\ref{app:eval}).

\textbf{Matched controlled comparisons.} With the same frozen \pizero{} policy
and the same Qwen3-VL-4B upper model, \sys{} scores $74.25\%$,
versus $20.3\%$ for PhyAgentOS and $6.0\%$ for Harness VLA
(Table~\ref{tab:main}). The gain over the frozen policy ($16.2\%$) is $58.0$
percentage points. PhyAgentOS with GPT-4o-mini reaches $21.3\%$;
the remaining Qwen3-VL-4B rows report our evaluated configurations. In a second
measurement of the reported agent ($74.1\%$), \sys{} alone wins $474$ paired
episodes and the policy alone $11$ (Appendix Fig.~\ref{fig:paired}).

\textbf{Published references and slow-brain replacement.} Harness VLA reports
$60.5\%$ (Codex) and $76.8\%$ (Claude Code) with \pizero{}-SFT, and Zetta reports
$71.1\%$. These are reference results under their published settings. Within \sys{}, replacing Qwen3-VL-4B with Claude Sonnet 5 through
a coding agent yields $80\%$ against $75\%$ over $40$ episodes on four untuned
cells, and $75\%$ with a tuned prompt (Fig.~\ref{fig:profiles}a).
Appendix~\ref{app:upper} details the scaffold and pilot latency.

\textbf{Per-task profile and breadth.} Gains are task-specific
(Fig.~\ref{fig:profiles}c--f): before the drawer changes, the agent succeeds on at
least $80\%$ of seeds in $27$ of $40$ tasks and on none in five, four of which
involve a drawer. On LIBERO-Plus, a separate configuration succeeds on $84.4\%$ of tasks,
with the lowest rates under camera and robot-state perturbations
(Fig.~\ref{fig:profiles}b). Appendix~\ref{app:eval} gives the evaluation protocols.

\textbf{Real-world experiments.}\label{sec:realworld}
On hardware (Fig.~\ref{fig:realrobot}), success reaches $90\%$ for circle
pick-and-place, $80\%$ for cup stacking, and $70\%$ for both bread
pick-and-place and drawer placement (Table~\ref{tab:realrobot}).
Each task is evaluated in $10$ trials with object positions varied across trials.
Appendix~\ref{app:realworld} describes the image-based
perception setup and full task instructions.

\subsection{Post-Selection Transfer}
\label{sec:postselection}

The $74.25\%$ result uses the adaptive development block.
With the system frozen before generating $800$ new states of the same cells,
\sys{} succeeds on $75.2\%$ versus $17.5\%$ for frozen
\pizero{}, a $57.8$-point gain with $473$ versus $11$ exclusive wins.
The four preregistered frozen snapshots rise from $60.9\%$ to $65.9\%$,
$74.2\%$ and $75.2\%$ (Table~\ref{tab:transfer_mechanism}A): the ranking is
preserved (Spearman $\rho=1.00$), and the $14.3$-point development gain becomes
$14.4$ points on new states. Of this transferred gain, $13.4$ points accrue by
stat39; the final $1.0$-point increment remains uncertain ($p=0.0768$).
Gains over the policy range from $52.5$ to $60.0$ points across suites, while 10-S remains lowest at
$66.5\%$. On $261$ untouched official states spanning $31/40$ cells,
\sys{} reaches $77.0\%$, versus $16.5\%$ for \pizero{};
Appendix~\ref{app:postselection} gives paired details.

\begin{figure}[t]
\centering
\includegraphics[width=\textwidth]{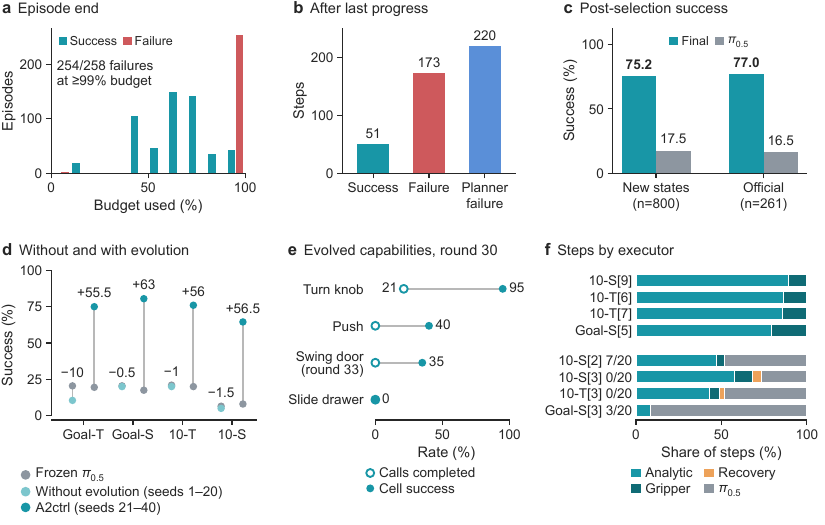}
\vspace{-5mm}
\caption{\textbf{Execution and transfer (top), physical harness (bottom).}
\textbf{a}~Budget used at episode end. \textbf{b}~Steps after the last progress.
\textbf{c}~Frozen champion versus policy: new states ($800$) and untouched official states ($261$).
\textbf{d}~Without and with evolution (A2ctrl on development seeds), each against the frozen policy on its own
seeds. \textbf{e}~Evolved capabilities; the frozen policy solves none of these
cells. \textbf{f}~Steps by executor: one filmed success per cell (top) and all
$20$ seeds of four hard cells (bottom).}
\label{fig:mech}
\end{figure}

\subsection{Fast-Slow Robot Agent}
\label{sec:fastresults}
\label{sec:fastslow}

\textbf{Preserving transient completion events.}\label{sec:fastevents}
In \emph{turn off the stove}, success held from control steps $44$--$55$, about
$0.55$\,s. Across $20$ replays, sampling every $20$ steps without latching succeeds on
$25\%$, versus $80\%$ for the bare policy and $95\%$ with latching and a
chunk-level read (Appendix Fig.~\ref{fig:suppl}a). Latching preserves transient success for termination. Appendix~\ref{app:ablation}
isolates latching at a fixed sampling period; Fig.~\ref{fig:timescales} reports
execution rates and measured latencies.

\textbf{Failures run out the budget.}\label{sec:episodes} In the analyzed round,
$254$ of the $258$ recorded failures ended at $99\%$ or more of their budget,
successes at a median of $67\%$ (Fig.~\ref{fig:mech}a), and across
$5{,}805$ episodes a failure spends $173$ steps after its last progress against
$51$ for a success, $220$ when the planner layer failed
(Fig.~\ref{fig:mech}b). These long suffixes identify where intervention could
save execution, but replacing the suffix yields $22$ alternative-only versus
$14$ control-only wins ($p=0.24$; Appendix~\ref{app:mechanism}). Detecting
unproductive execution and supplying an effective replacement action are
therefore separate requirements. An exploratory learned fast brain also scores $74.1\%$
(Appendix~\ref{app:learned}).\label{sec:learned}

\subsection{Dynamic Physical Harness}
\label{sec:harness}

\textbf{The initial harness.} On seeds $1$--$20$, the harness without evolution
succeeds on $13.9\%$ versus $17.1\%$ for the frozen policy (Fig.~\ref{fig:mech}d):
$9$ of $40$ cells worsen and $1$ improves (Fig.~\ref{fig:percell}); half the loss
relates to the stove window. Development added capabilities and changed execution
mechanisms.

\textbf{Capability composition in recorded episodes.}\label{sec:reliability}
The records show successful episodes despite individual capability failures
(Fig.~\ref{fig:mech}e). \code{turn\_knob\_object} completes $21\%$ of its
invocations while its source cell improves from $0\%$ to $95\%$. Refusing
\code{push\_object}, for which the jaw is too narrow, in favor of carrying the
plate illustrates a substitution in a cell that improves from $0\%$ to $40\%$
(Fig.~\ref{fig:example}). A local refusal can thus lead to a feasible alternative;
invocation-level completion alone misses this episode-level benefit.
The placement goal is preserved while the physical operation changes: the library
supplies the alternative, and execution governance determines when to select it.
\textbf{Capability usage.}\label{sec:vlausage}
Of $770$ archived control episodes, $570$ never invoke the VLA, with $96.7\%$
success. Analytic execution supplies most successful behavior; policy
calls concentrate in difficult episodes after analytic execution struggles
(Appendix~\ref{app:mechanism}). In retained failures, VLA execution reaches
$91.4\%$ of steps in the hardest cells, where no cell exceeds $35\%$
(Fig.~\ref{fig:mech}f; Appendix~\ref{app:retry}).

\begin{figure}[t]
\centering
\includegraphics[width=\textwidth]{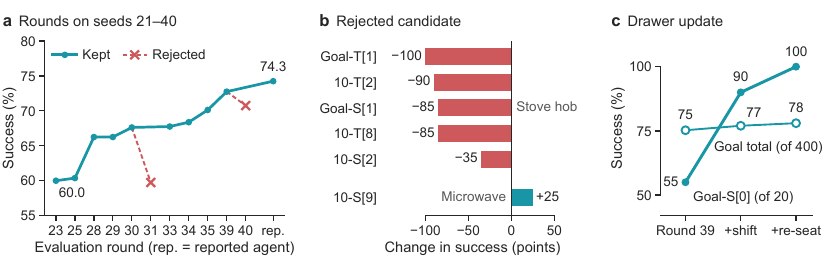}
\vspace{-5mm}
\caption{\textbf{Development history.} \textbf{a}~Full evaluation of
each round on seeds $21$--$40$. \textbf{b}~Change from round $30$ with the
rejected candidate, on the hob and microwave cells. \textbf{c}~Drawer update.}
\label{fig:evolution}
\vspace{2mm}
\includegraphics[width=\textwidth]{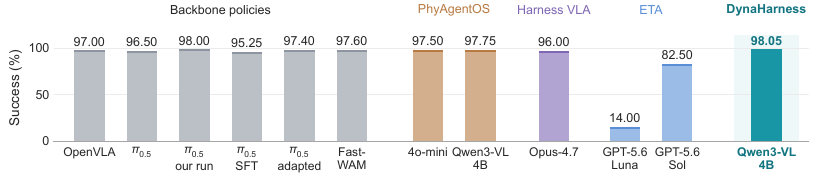}
\vspace{-5mm}
\caption{\textbf{Standard LIBERO success.} Means over four suites; each bar
matches one row of Appendix Table~\ref{tab:libero}. \sys{} uses settings adapted
to standard LIBERO. Labels identify upper models for agentic baselines.}
\label{fig:libero}
\end{figure}

\subsection{Attribution-Driven Self-Evolution}
\label{sec:evoresults}

\textbf{Changes across development rounds.} Across rounds, development success
rose from $60.0\%$ to $74.25\%$, and two rounds were not kept
(Fig.~\ref{fig:evolution}a). In the analyzed round, $27$ of $40$ cells gain five
or more successes over the frozen policy, $3$ lose a few and $7$ remain at zero
(Appendix Fig.~\ref{fig:percell}). Among $207$ failures in the second measurement,
$97$ involve a drawer, $43$ a stove and $30$ a microwave. This concentration makes
shared capability failures a concrete target for updates across multiple episodes.

\textbf{A capability update following attribution.}\label{sec:drawer} Attribution charged
the drawer cells to the capability layer. A targeted probe of
\code{goal\_swap[0]} found three physical causes: the forearm wedged against the
wine rack, the closure limit rejected about a third of real grasps on the
$15.4$\,mm handle, and the hand often sat too high after contact. Two changes
stated in these quantities raised cell success from $55\%$ to $90\%$ and $100\%$
over $20$ seeds, and Goal success from $75.25\%$ to $78.0\%$ over $400$ episodes (Fig.~\ref{fig:evolution}c).
The first was an intermediate development baseline (Appendix~\ref{app:drawer}).

\textbf{Broader validation after the paired gate.}\label{sec:gate} A candidate
that passed its paired gate lost $79$ successes on five stove-hob cells in the
full round (Fig.~\ref{fig:evolution}b); reverting it restored them. Pairing controls
initial-state variation, while broader validation tests whether a local fix
changes behavior in other evaluated task cells. A second candidate was not kept after
a lower full-round score (Appendix~\ref{app:twostage}).
Retaining an update in the shared library therefore requires validation beyond
the task cells that motivated it.
The paired gate tests the targeted change, while broader evaluation checks
its interaction with capabilities already used elsewhere in the library.

\subsection{What Makes the Physical Harness Work?}
\label{sec:causal}

\textbf{Capability contribution.} Removing seven analytic contact skills reduces
success from $74.0\%$ to $16.6\%$, near bare
\pizero{}'s $16.2\%$. Recovery/intervention removal yields
$73.1\%$ ($p=0.21$); removing both groups yields $15.8\%$
(Table~\ref{tab:transfer_mechanism}B). Analytic skills supply the main task competence.

\textbf{Three executors, one library.} On the same $800$ development states,
A2seq freezes an initial skill sequence ($63.75\%$). A2static retains Full's
one-step planner interface and queries updated observations after successful
nominal skills ($63.88\%$), but disables failure-triggered replanning,
substitution, reordering, recovery insertion and verifier-conditioned branching.
Full retains these dynamic responses ($74.0\%$). A2seq also changes
planning horizon and output schema; A2static is the primary control
(Appendix~\ref{app:static}).

Full exceeds A2static by $10.125$ points ($89/8$ exclusive wins,
cell-bootstrap $95\%$ CI $[3.375,18.250]$, $p=2.00\times10^{-18}$).
A2static exceeds A2seq by only $0.125$ points ($11/10$ exclusive wins, $p=1$).
This joint gain extends beyond nominal replanning and exceeds the
recovery-only effect.
A2static already observes state changes between successful skills; Full also
responds to failures and verifier feedback, adapting how the available
capabilities are used.

\textbf{Mechanism diagnostics.} Across $783$ trace-matched Full/A2static pairs,
Full records $566$ failure/escalation replans, $471$ substitutions and $141$
recovery insertions, versus zero for A2static. A2static instead uses $534$ fixed
step retries and $255$ plan reexecutions. These descriptive counts confirm
execution-policy differences, not individual causal effects
(Tables~\ref{tab:mechanism_counts}--\ref{tab:routing_new}). Latching and capability ablations appear in Appendix~\ref{app:ablation}.

\section{Conclusion}

\sys{} governs analytic skills, recovery skills and a frozen policy through a
dynamic physical execution substrate. Failure evidence localizes faults and
directs reusable capability or mechanism revisions, admitted through paired
regression checks. The frozen system reaches $75.2\%$ on new initial states of
the same cells versus $17.5\%$ for frozen \pizero{}; snapshot gains transfer from
$14.3$ points during development to $14.4$ on new states. Analytic skills supply
the main task competence: removing them lowers success from $74.0\%$ to $16.6\%$,
while closed-loop execution with the
same library improves over nominal one-step replanning by $10.125$ points.

\par
\clearpage
\setlength{\parskip}{\mainSavedParskip}
\setlength{\textfloatsep}{\mainSavedTextfloatsep}
\setlength{\floatsep}{\mainSavedFloatsep}
\setlength{\intextsep}{\mainSavedIntextsep}
\flushbottom

\IfFileExists{sec_statements.tex}{

\subsection*{AI use statement}
We used generative AI tools, including Claude (Anthropic) and ChatGPT (OpenAI), for limited language assistance during the preparation of this paper, primarily to improve the clarity, grammar, and phrasing of selected passages. Generative AI was not used to propose hypotheses, design experiments, implement \sys{} or the baselines, conduct experiments, analyze results, or draw scientific conclusions. All AI-assisted edits were reviewed and revised by the authors, who take full responsibility for the content, claims, and conclusions of this work. Models that form part of the evaluated systems are described separately as components of the experiments in Section~\ref{sec:setup}.

\subsection*{Reproducibility statement}
Section~\ref{sec:setup} and Appendices~\ref{app:protocols} and~\ref{app:eval}
specify the benchmarks, evaluation cells, seed blocks, step budgets, frozen
checkpoints, planner models, and the protocol and source of each baseline result.
Section~\ref{sec:method} and Appendix~\ref{app:method} describe the execution
contract and admission rule (Eq.\ref{eq:gate}). Appendix~\ref{sec:limitations}
states which experiments retain per-episode records. We additionally provide
an anonymized supplementary package containing the implementation of
\sys{}, evaluation configurations and scripts, and the run records used to
reproduce the reported aggregate results and paired comparisons.}{}

\edef\savedReferenceBibsep{\the\bibsep}
\setlength{\bibsep}{2pt}

\setlength{\bibsep}{\savedReferenceBibsep}

\newpage
\appendix
\raggedbottom
\setlength{\parskip}{3pt}
\ifdefined\arxivpreprint
  \setlength{\parskip}{2pt}
\fi
\setlength{\textfloatsep}{12pt plus 2pt minus 2pt}
\setlength{\floatsep}{10pt plus 2pt minus 2pt}
\setlength{\intextsep}{10pt plus 2pt minus 2pt}
\setcounter{topnumber}{4}
\setcounter{totalnumber}{6}
\renewcommand{\floatpagefraction}{0.8}

\section*{Appendix Contents}
\vspace{-1mm}
\noindent
\begin{tabular}{@{}p{0.92\textwidth}r@{}}
\hyperref[app:method]{\ref*{app:method}\quad Method Details} & \pageref{app:method} \\
\hyperref[app:protocols]{\ref*{app:protocols}\quad Experimental Setup Details} & \pageref{app:protocols} \\
\hyperref[app:eval]{\ref*{app:eval}\quad Evaluation Details} & \pageref{app:eval} \\
\hyperref[app:results]{\ref*{app:results}\quad Additional Results} & \pageref{app:results} \\
\hyperref[app:cases]{\ref*{app:cases}\quad Case Studies} & \pageref{app:cases} \\
\hyperref[app:pertask]{\ref*{app:pertask}\quad Per-Task Results} & \pageref{app:pertask} \\
\hyperref[sec:ranbaselines]{\ref*{sec:ranbaselines}\quad Baseline Records} & \pageref{sec:ranbaselines} \\
\hyperref[app:twostage]{\ref*{app:twostage}\quad Two-Stage Admission of a Recovery Candidate} & \pageref{app:twostage} \\
\hyperref[sec:limitations]{\ref*{sec:limitations}\quad Limitations} & \pageref{sec:limitations} \\
\end{tabular}

\section{Method Details}
\label{app:method}

This section gives the parts of Section~\ref{sec:method} that the main text
states briefly.

\paragraph{Planning and completion interfaces.}
The arguments $\alpha_j$ in Eq.~\ref{eq:ctx} identify scene entities and goal
relations. The fast brain resolves geometry and enforces refusal and termination.
In LIBERO, the execution interface obtains completion from the benchmark task
predicate. It records detected events independently of slow-brain calls, and the
fast brain reads the accumulated verdict at its next decision. Equation~\ref{eq:verify}
separates these event checks from the $2$\,Hz decision schedule: a Boolean OR
retains detected events but cannot recover events missed by the underlying
checks. This task verdict is distinct from a capability's local completion status.

\textbf{Retry, recovery and switching.} When a command ends without success,
the next decision belongs to the fast brain. It can ground the same step again,
substitute another eligible capability, invoke a recovery skill, or hand the
segment to \code{vla\_act}, and each choice is written to the evidence store with
its reason. A refusal, or a step for which no eligible capability remains, is
returned to the slow brain for a new plan.
Its decisions and their outcomes are logged in a form a learned fast brain could
be trained on; Appendix~\ref{app:learned} reports what training one produced.

\paragraph{The physical execution contract.}
\textbf{Grounding records.} The execution interface supplies geometric inputs
for capability parameters and preconditions. In the evaluated LIBERO setup,
these inputs come from simulator state; on hardware, scene information comes
from camera images (Appendix~\ref{app:realworld}). The evidence store records
grounding outcomes and refusal reasons. The benchmark completion predicate
supplies the LIBERO task verdict separately from geometric grounding.

\textbf{Budget, lease and refusal.} A command carries a step budget, a lease and
a velocity envelope of $2$\,rad/s. The budget is the binding constraint:
LIBERO-Goal allows $300$ environment steps, one pick-and-place costs about $230$
and a single knob turn has a median cost of $211.5$, so the two do not both fit
in one episode. A capability that is unable to afford its own completion refuses
before it starts rather than spending the episode discovering the same fact, and
because the refusal is a value, its reason is written down and can later be
attributed.

\textbf{Latched completion verdict.} Once true, the verdict stays true
(Eq.~\ref{eq:verify}) and is recorded with the command it ended, which separates
completed commands from those that ran out of budget.

\textbf{Evidence store.} Each episode has an append-only store in which every
observation, decision and outcome is written with the reason it was made. The
same store supplies the history $h_j$ of Eq.~\ref{eq:ctx}, the input to failure
attribution, and the record self-evolution is judged against.

\textbf{Safety and the robot seam.} An envelope, lease and budget check runs at
$50$\,Hz underneath both brains. The harness issues bounded, verifiable commands
and does not run a servo loop, since control at $100$ to $1000$\,Hz belongs to the
robot controller; simulated, MuJoCo workcell and remote adapters sit behind the
same interface.
Table~\ref{tab:stepcosts} gives the step cost of common commands.

\begin{table}[t]
\centering
\caption{\textbf{Steps per command stage}, from the retained event stores.}
\label{tab:stepcosts}
\footnotesize
\setlength{\tabcolsep}{5pt}
\begin{tabular}{@{}lrl@{}}
\toprule
Command & Steps & Note \\
\midrule
Approach & 30 & \\
Guarded descent & about 30 & 41 before the speed work \\
Close the jaw & 10 & $0.5$\,s gripper command; 20 at $1.0$\,s \\
Lift & 23 & \\
Carry over the corridor & 60--80 & depends on the corridor height \\
Lower & 15 & \\
Release & 10 & \\
Retreat & 23 & \\
One pick-and-place & about 230 & \\
Top-drawer pull & 239 & 297 before the wrist ramp \\
Wrist turn pair & 67 & 108--115 before the four-step ramp \\
\bottomrule
\end{tabular}
\end{table}

\paragraph{Analytic and learned capabilities.} The library offers geometric and learned
execution routes. Analytic skills own the
stages that measured geometry determines: approach, guarded descent, transport
over a corridor and placement, computed from execution-interface geometry
with grounding checks before motion. Recovery skills return the arm to a pose recorded earlier in the
episode or release and retreat. The VLA is called for contact whose dynamics
that geometry does not determine. Treating it as one capability rather than as
the controller is what lets the fast brain stop it when the verdict latches and
hold it to the same budget as an analytic skill.

When no eligible capability grounds against the scene, or every installed route
has spent its budget, the remaining steps fall to \code{vla\_act}. That outcome
is recorded as the reason the policy was left to own the episode, and it is one
of the signals failure attribution reads. Section~\ref{sec:vlausage} reports this routing behavior in retained traces.

\paragraph{Diagnostic granularity and ordered attribution.}
The hyperparameter $N$ in Eq.~\ref{eq:attrib} specifies the resolution of the
diagnostic label space. It controls how finely execution evidence is partitioned
into categories for candidate revision.
A coarse partition can merge failures requiring different revisions; a finer
partition requires evidence that distinguishes the additional categories.
The relevant criterion is thus whether the command record supports a distinction
that changes the candidate repair target. For example, unresolved grounding and
unsuccessful execution after admission motivate different revisions even when
both episodes end in task failure.

The evaluated configuration uses $N=13$, spanning context and planning,
grounding, dispatch, capability execution, verification, recovery and
infrastructure. Let $d_j(E_i)$ denote the diagnostic check for layer $j$.
For an episode with at least one matching check, ordered attribution implements
$A_N(E_i)=\min\{j\in\{1,\ldots,N\}:d_j(E_i)=1\}$.
The ordering prioritizes upstream conditions and resolves simultaneous matches;
an episode without a matching check is reported as an error.
The resulting label is a diagnostic hypothesis for revision, not an identified
causal effect. Here, $N=13$ is a configuration choice; the reported experiments
do not establish an optimal granularity or sensitivity to $N$.

\begin{figure}[t]
\centering
\includegraphics[width=\textwidth]{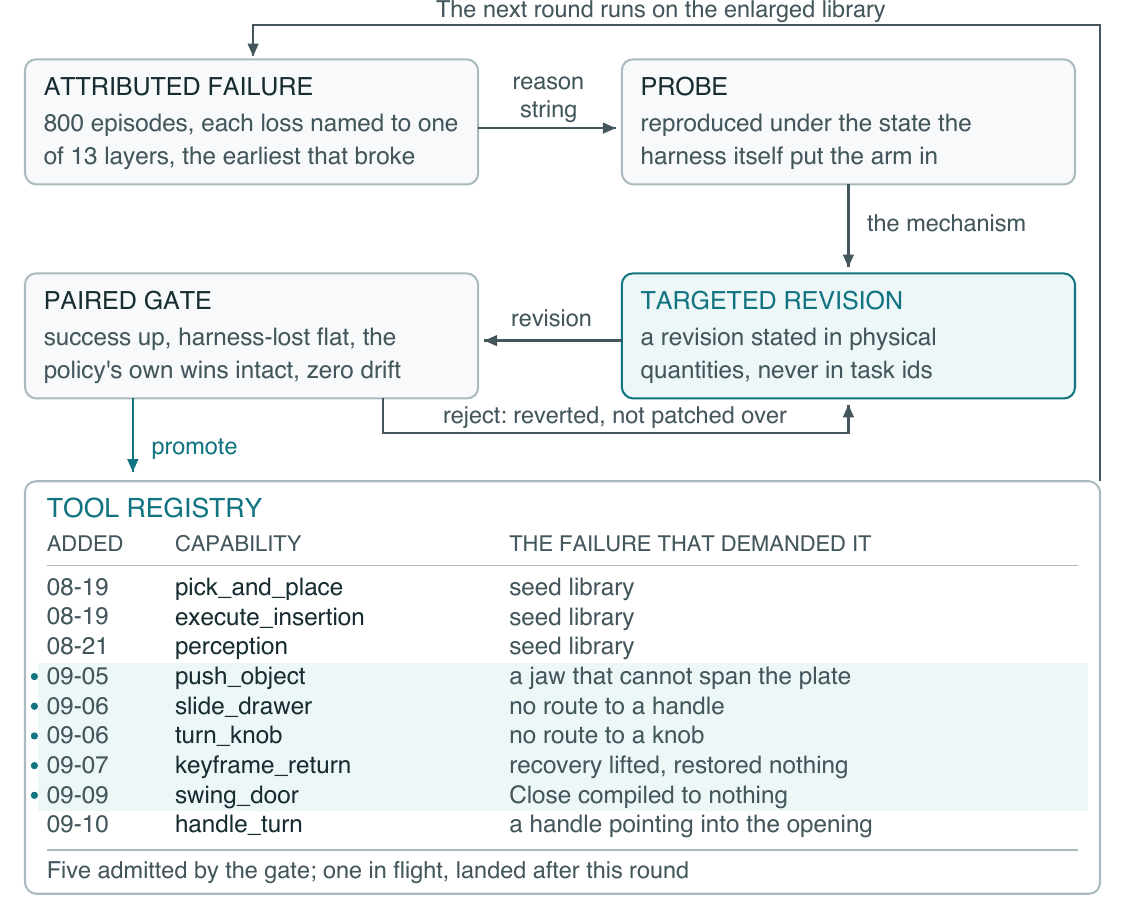}
\caption{\textbf{The evolution loop and its registry.} Registry entries track capability additions following fault probing, targeted revision and paired validation.}
\label{fig:evoloop}
\end{figure}

\paragraph{The evolution loop.} Figure~\ref{fig:evoloop} shows the loop and the
registry it produced. The loop connects failure attribution, fault probing and
targeted revision with paired validation and registry admission.

\textbf{Probe.} The named fault is reproduced from the state the
harness itself put the arm in rather than from a reset. A placement failure that
exists only once a mug is already on a plate and the corridor above it is
occupied does not appear in a fresh scene.

\textbf{Revise.} The revision is formulated in physical
quantities, such as a hinge radius, a hand span or a
ramp length, rather than as a branch keyed to a task identifier. These parameters expose the geometric conditions under which the revision
is intended to apply.

\textbf{Registry.} An admitted change enters $\mathcal{L}$ and the
next round runs on it; a rejected one is reverted rather than patched over.
Section~\ref{sec:evoresults} and Appendix~\ref{app:twostage} report candidates
that passed local screening but were not kept after the broader round. The evidence stores that drive this loop also hold every fast-brain
decision with its outcome, which is the data a learned fast brain would be
trained on (Appendix~\ref{app:learned}).

\section{Experimental Setup Details}
\label{app:protocols}

\paragraph{Matched controlled comparisons.} Our main LIBERO-Pro comparisons
use the same four task cells and benchmark success metric. \sys{}, PhyAgentOS
and Harness VLA are evaluated with the same Qwen3-VL-4B upper model, frozen
public \pizero{} LIBERO checkpoint, evaluation budget and episode protocol:
$10$ tasks per cell and $20$ episodes per task, $800$ episodes in total.
Success is determined by the benchmark predicate over the full denominator.
\sys{} and its frozen-policy control use development seeds $21$--$40$;
Section~\ref{sec:setup} and Appendix~\ref{app:eval} describe our evaluation.
Table~\ref{tab:protocols} summarizes the models and evaluation sizes of our runs;
Table~\ref{tab:provenance} records each result's source.

Step budgets are each benchmark's own: $220$, $280$,
$300$ and $520$ environment steps for the Spatial, Object, Goal and LIBERO-10
suites, kept by LIBERO-Pro.

\begin{table}[ht]
\centering
\caption{\textbf{Models and evaluation settings of our LIBERO-Pro runs.}}
\label{tab:protocols}
\scriptsize
\setlength{\tabcolsep}{3pt}
\renewcommand{\arraystretch}{1.1}
\begin{tabular}{@{}>{\raggedright\arraybackslash}p{0.24\textwidth}>{\raggedright\arraybackslash}p{0.21\textwidth}>{\raggedright\arraybackslash}p{0.24\textwidth}>{\raggedright\arraybackslash}p{0.24\textwidth}@{}}
\toprule
System & Upper model & Low-level policy & Evaluation setting \\
\midrule
\sys{} & Qwen3-VL-4B & frozen \pizero{} & four cells, $200$ episodes each \\
\pizero{} (our run) & none & frozen \pizero{} & four cells, $200$ episodes each \\
PhyAgentOS & Qwen3-VL-4B; GPT-4o-mini & frozen \pizero{} & four cells, $200$ episodes each \\
Harness VLA & Qwen3-VL-4B & frozen \pizero{} & four cells, $200$ episodes each \\
ENPIRE & Qwen3-VL-4B & \pizero{} via generated code & four cells, $200$ episodes each \\
Zetta & Qwen3-VL-4B & \pizero{} via tool interface & four cells, $200$ episodes each \\
ASPIRE & Qwen3-VL-4B & generated code & four-cell campaign; no episode executed \\
CaP-Agent0 & Qwen3-VL-4B & generated code and manipulation tools & four cells, $10$ episodes each; pointing model disabled \\
EmbodiedSkills & Qwen3-VL-4B & not specified & four-cell campaign \\
\bottomrule
\end{tabular}
\end{table}

\begin{table}[ht]
\centering
\caption{\textbf{Source of every row of Tables~\ref{tab:main} and~\ref{tab:libero}}, checked against the cited table.}
\label{tab:provenance}
\scriptsize
\setlength{\tabcolsep}{3pt}
\renewcommand{\arraystretch}{1.1}
\begin{tabular}{@{}>{\raggedright\arraybackslash}p{0.16\textwidth}>{\raggedright\arraybackslash}p{0.22\textwidth}>{\raggedright\arraybackslash}p{0.56\textwidth}@{}}
\toprule
Row & Source & Location and note \\
\midrule
OpenVLA, \pizero{} & \citet{liberopro2025} & Tables 2 and 4, Average rows (Task and Pos columns); 50 episodes per task \\
\pizero{} (our run) & ours & frozen \pizero{} LIBERO checkpoint, seeds $21$--$40$, all episode rows retained \\
\pizero{}-SFT & \citet{harnessvla2026} & Table 3, row $\pi_{\mathrm{RLinf}}$ (RLinf \code{pi05\_libero130\_fullshot}) \\
Fast-WAM & authors; \citet{fastwam} & LIBERO-Pro cells supplied by the authors; LIBERO from Table 2 of the publication \\
CaP-Agent0 & \citet{capx}; \citet{aspire2026} & Goal cells: CaP-X Table 7 Average ($0.168$, $0.256$); 10-T and 10-S: ASPIRE Table 5 ($0.024$, $0.052$) \\
RHO & \citet{rho} & Table 7 Average, RHO columns ($0.556$, $0.506$) \\
SPARK & \citet{spark} & Table 1, row Spark, Adaptive \\
Pigey & \citet{pigey} & Table 15, one row per reasoner (all nine in Table~\ref{tab:pigey}) \\
VLS & \citet{vls} & Table I, row \pizero{} (LeRobot) + VLS \\
PhyAgentOS, Zetta, ENPIRE with Qwen3-VL-4B or GPT-4o-mini & ours & run summaries of the second campaign, $200$ episodes per cell (\code{aspire\_phyagent.xlsx}, \code{ENPIRE\_LIBERO\_Pro\_success\_rates}); the earlier $400$-episode PhyAgentOS run and its per-episode records are in Appendix~\ref{sec:ranbaselines} \\
Harness VLA & \citet{harnessvla2026} & Table 3, rows Harness VLA (Codex) and (CC); Avg.\ over these four of its eight cells \\
Harness VLA with Qwen3-VL-4B & ours & campaign summary, $200$ episodes per cell ($6.0\%$ overall; suite success $14.0\%$, $7.0\%$, $3.0\%$, $0.0\%$); per-episode rows were not retained \\
Zetta & \citet{zetta2026} & Table 3 \\
ASPIRE & \citet{aspire2026} & Goal: Table 2 ($0.45$, $0.81$); 10-T and 10-S: Table 5, $N = 90$, zero-shot ($0.383$, $0.226$) \\
EmbodiedSkills with Qwen3-VL-4B & ours & our Qwen3-VL-4B run over the four cells; author-confirmed results \\
\sys{} & ours & archived final aggregate, seeds $21$--$40$: $74.25\%$; author-confirmed suite rates. Historical paired analyses use a separate $74.1\%$ remeasurement \\
\midrule
\multicolumn{3}{@{}l}{\textit{Table~\ref{tab:libero} (LIBERO)}}\\
OpenVLA, \pizero{} & \citet{liberopro2025} & Tables 2 to 5, unperturbed (Ori) columns \\
\pizero{} (our run), PhyAgentOS, \sys{} & ours & run records (\sys{}: LIBERO configuration in Table~\ref{tab:liberoconfigs}) \\
\pizero{}-SFT, Harness VLA & \citet{harnessvla2026} & Table 2 \\
\pizero{}, task-adapted & \citet{embodiedskills2026} & Table 3; the paper credits these numbers to its task-adapted low-level policy \\
Fast-WAM & \citet{fastwam} & Table 2 \\
ETA & \citet{eta2026} & Luna: Table 2, 40 tasks $\times$ 10 seeds; Sol: Table 8, Pass@1 counts of 10 tasks per suite \\
\bottomrule
\end{tabular}
\end{table}

\paragraph{Sources and table conventions.} We ran \sys{}, the frozen-policy rows
labeled ``our run'', all Qwen3-VL-4B harness rows, and PhyAgentOS with GPT-4o-mini;
Table~\ref{tab:provenance} identifies their sources. Other rows are external
reported references from the corresponding papers, with the author-supplied
entries noted below. A harness with several upper models appears once per model;
``--'' denotes an unreported cell, and the highest and second-highest values in
each column are bold and underlined, respectively, including ties.
The Fast-WAM LIBERO-Pro cells and Zetta's upper-model identity are supplied by
the authors of this paper. The published Harness VLA rows use RLinf
\pizero{}-SFT; the task-adapted \pizero{} row in Table~\ref{tab:libero} is the
low-level policy reported by EmbodiedSkills. Table~\ref{tab:pigey} gives Pigey's
full reasoner sweep. The matched comparison controls the listed model, policy
and evaluation settings; candidate-development effort and skill libraries are
system-specific. ASPIRE with Qwen3-VL-4B executed no episodes, and our
CaP-Agent0 adaptation disabled its pointing model. Both are documented as
diagnostic runs in Appendix~\ref{sec:ranbaselines}, outside the main success table.

\begin{table}[ht]
\centering
\caption{\textbf{Pigey with nine reasoners} over the same frozen \pizero{} (Table~15 of \citet{pigey}), 10 trials per task.}
\label{tab:pigey}
\scriptsize
\setlength{\tabcolsep}{5pt}
\begin{tabular}{@{}lcc|lcc@{}}
\toprule
Reasoner & Goal-T & Goal-S & Reasoner & Goal-T & Goal-S \\
\midrule
GPT-5.5 (low)      & 28 & 42 & Gemini 3.1 Pro    & 22 & 44 \\
GPT-5.5 (medium)   & 24 & 44 & Claude Haiku 4.5  & 20 & 38 \\
GPT-5.5 (high)     & 26 & 38 & Claude Sonnet 4.6 & 30 & 42 \\
Gemini Rob-ER 1.6  & 34 & 44 & Claude Opus 4.7   & 22 & 44 \\
Gemini 3.5 Flash   & 28 & 48 &                   &    &    \\
\bottomrule
\end{tabular}
\end{table}

\paragraph{Metrics and evaluation details.} The primary metric is success rate
over all episodes, with Wilson $95\%$ intervals where they matter; paired
comparisons report discordant pairs and an exact test. The \emph{step share} of
an episode is the fraction of its environment steps spent under analytic skills
and under the VLA, and the \emph{conversion} of a capability is its source cell's
success count before and after admission on the same seeds. Episodes lost to evaluation infrastructure count as failures.
Outcome counts and run-specific processing are given below; the capability
ablation diagnostics are retained in Appendix~\ref{app:bb2}. Appendix~\ref{app:eval} gives the outcome composition, the evaluation
history of the agent and the LIBERO-Plus protocol, and
Appendix~\ref{sec:ranbaselines} the baseline records. The results of
Section~\ref{sec:drawer} and Appendix~\ref{app:learned} are taken from the run reports
and dataset manifests of the machines that produced them.

\subsection{Real-World Platform and Task Instructions}
\label{app:realworld}

\begin{figure}[!htb]
  \centering
  \includegraphics[width=\textwidth]{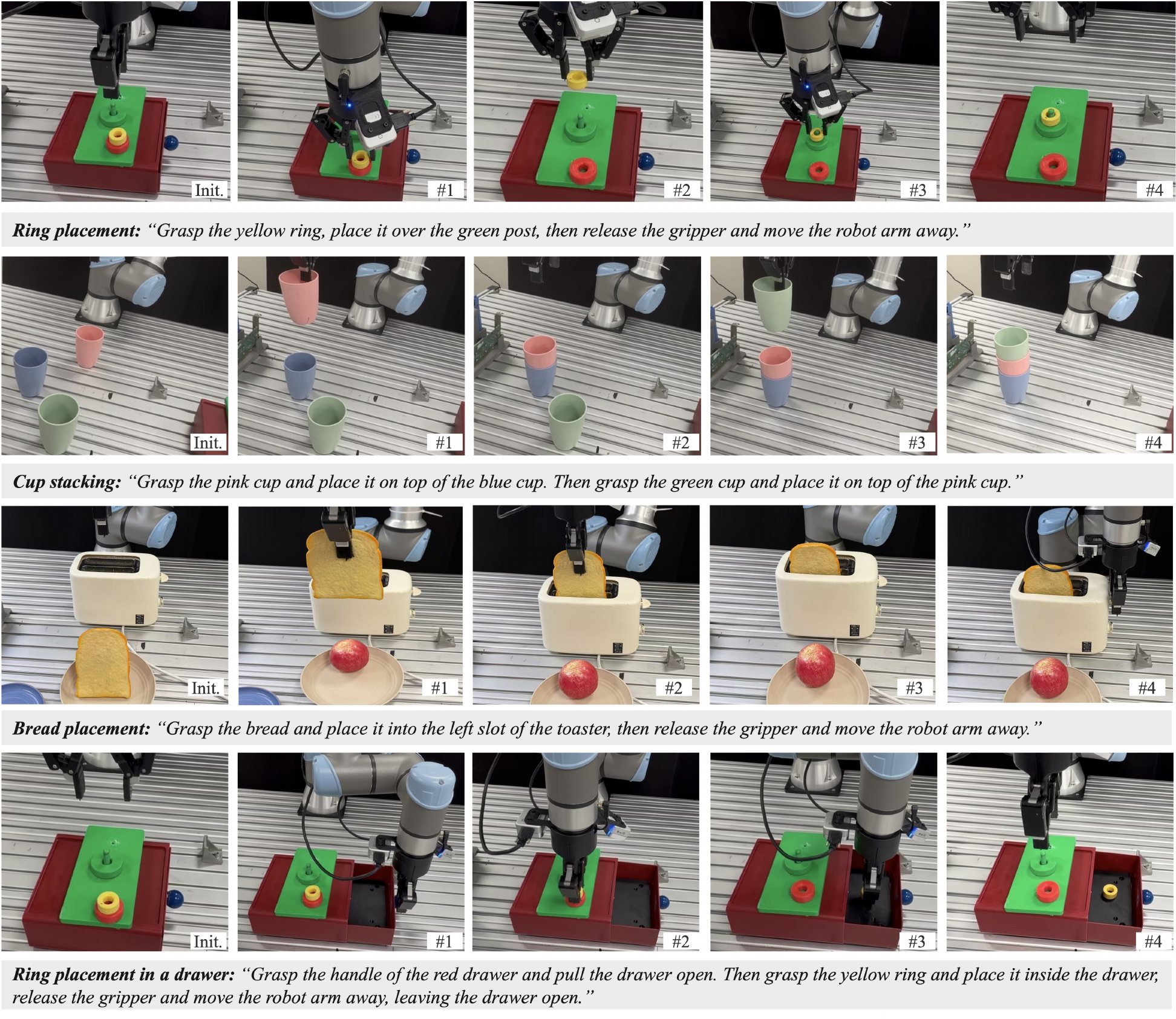}
  \caption{Real-world execution sequences for ring placement, cup stacking,
  bread placement, and ring placement in a drawer (top to bottom). Each row
  shows the initial scene followed by four execution stages, with the
  corresponding language instruction below.}
  \label{fig:realworld_sequences}
\end{figure}

\paragraph{Platform and visual input.}
The physical platform uses a Qwen 4B language model, the SAM3 visual
segmentation model, a UR7e robot arm, and Intel RealSense D435 and D405 cameras
(Fig.~\ref{fig:realrobot}). Scene information is obtained from images captured
by these cameras, with SAM3 providing visual segmentation.
Table~\ref{tab:realrobot} reports success on the four tasks below.
Figure~\ref{fig:realworld_sequences} illustrates their execution on the physical platform.

\paragraph{Trial protocol.}
Each task is evaluated in $10$ trials, for $40$ trials in total. Object placements
vary across trials to assess performance under position changes. Success rates
are $90\%$ for ring placement, $80\%$ for cup stacking, and $70\%$ each for bread
and drawer placement, as reported in
Table~\ref{tab:realrobot}.

\paragraph{Complete task instructions.}
The complete instructions are given below in English translation.
\begin{enumerate}
\setlength{\itemsep}{2pt}
\setlength{\parsep}{0pt}
\item \textbf{Ring placement.} ``Grasp the yellow ring, place it over the green
post, then release the gripper and move the robot arm away.''
\item \textbf{Cup stacking.} ``Grasp the pink cup and place it on top of the blue cup. Then grasp the green cup and place it on top of the pink cup. ''
\item \textbf{Bread placement.} ``Grasp the bread and place it into the left
slot of the toaster, then release the gripper and move the robot arm away.''
\item \textbf{Ring placement in a drawer.} ``Grasp the handle of the red drawer
and pull the drawer open. Then grasp the yellow ring and place it inside the
drawer, release the gripper and move the robot arm away, leaving the drawer
open.''
\end{enumerate}

\section{Evaluation Details}
\label{app:eval}

\begin{figure}[t]
\centering
\includegraphics[width=0.92\textwidth]{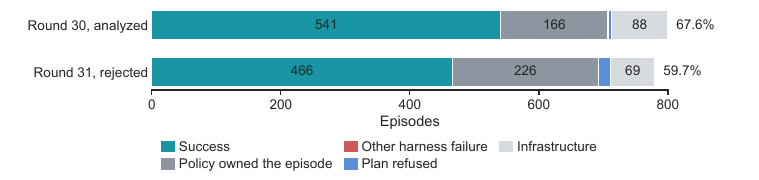}
\caption{\textbf{Outcome composition with and without the rejected candidate.}
Round $31$ ran $780$ episodes; percentages are of each round's episodes.}
\label{fig:composition}
\end{figure}

\paragraph{Outcome composition and denominator.} Table~\ref{tab:denominator}
assigns every episode of the analyzed round to an outcome, and
Fig.~\ref{fig:composition} compares it with the round that carried the rejected
candidate of Section~\ref{sec:gate}. Of the $800$ episodes, $88$ were lost to a
workcell connection that refused or dropped, which is not a decision the agent
made. Every success rate in the paper counts them as failures; excluding them,
the analyzed round succeeds on $76.0\%$ over $712$ episodes, compared with
$67.6\%$ over all $800$ episodes.

\paragraph{Evaluation history.} Table~\ref{tab:variants} gives each arm on both
seed blocks, and Table~\ref{tab:rounds} lists every full
LIBERO-Pro evaluation of the agent on the development block, including the two
whose changes were not kept. The analyzed round is \code{stat30}. The agent of
Table~\ref{tab:main} is \code{stat39} with the two drawer changes of
Section~\ref{sec:drawer}; the reported development aggregate is $74.25\%$, while a complete paired
remeasurement scores $74.1\%$. Table~\ref{tab:main} retains the former;
historical paired analyses use the latter's episode records. The new concurrent
mechanism-ablation control scores $74.0\%$, and post-selection results use
the separate state banks in Appendix~\ref{app:postselection}.
The latched verdict of Section~\ref{sec:fastevents} entered the agent in
\code{stat23}.

\paragraph{Unperturbed LIBERO.}
\sys{} achieves $98.05\%$ with sampler settings adapted on six
of the $40$ standard LIBERO tasks (Table~\ref{tab:libero}); the policy reference
scores $98.0\%$. Table~\ref{tab:liberoconfigs} records these configurations
alongside the frozen LIBERO-Pro champion. The latter's archived run scores
$46.7\%$, including $1064$ tick-limit failures under a $600$-tick
configuration: $69$, $348$, $149$ and $498$ across Spatial, Object, Goal and Long.
These execution limits confound interpretation of the frozen run as a measure
of transfer to standard LIBERO.

\begin{table}[t]
\centering
\captionof{table}{\textbf{Success rate (\%) on standard LIBERO.}}
\label{tab:libero}
\centering
\scriptsize
\setlength{\tabcolsep}{3.5pt}
\renewcommand{\arraystretch}{1.33}
\begin{tabular}{@{}ll*{4}{S[table-format=3.1]}S[table-format=2.2]@{}}
\toprule
Method & Upper VLM & {Spatial} & {Object} & {Goal} & {Long} & {Avg.} \\
\midrule
\multicolumn{7}{@{}l}{\textit{Backbone policies}}\\
OpenVLA              & none & 98.0 & 99.0 & 98.0 & 93.0 & 97.00 \\
\pizero{}            & none & 98.0 & 98.0 & 97.0 & 93.0 & 96.50 \\
\pizero{} (our run)                        & none & {\underline{99.0}} & 99.0 & {\textbf{98.6}} & 95.4 & {\underline{98.00}} \\
\pizero{}-SFT       & none & {\underline{99.0}} & 96.0 & 97.0 & 89.0 & 95.25 \\
\pizero{}, task-adapted & none & {\underline{99.0}} & 98.6 & {\underline{98.4}} & 93.6 & 97.40 \\
Fast-WAM                   & none & 98.2 & {\textbf{100.0}} & 97.0 & 95.2 & 97.60 \\
\addlinespace[3pt]
\multicolumn{7}{@{}l}{\textit{Agentic baselines}}\\
PhyAgentOS                       & GPT-4o-mini & {\textbf{100.0}} & {\textbf{100.0}} & 98.0 & 92.0 & 97.50 \\
PhyAgentOS                       & Qwen3-VL-4B & 98.0 & 99.0 & 98.0 & {\underline{96.0}} & 97.75 \\
Harness VLA         & Claude Opus-4.7 & 97.0 & {\textbf{100.0}} & 94.0 & 93.0 & 96.00 \\
ETA                  & GPT-5.6 Luna & 8.0 & 26.0 & 21.0 & 1.0 & 14.00 \\
ETA                  & GPT-5.6 Sol & {\textbf{100.0}} & 80.0 & 80.0 & 70.0 & 82.50 \\
\midrule
\sys{} & Qwen3-VL-4B & 98.8 & {\underline{99.6}} & 97.0 & {\textbf{96.8}} & {\textbf{98.05}} \\
\bottomrule
\end{tabular}
\par\vspace{2pt}
\parbox{\linewidth}{\scriptsize Bold/underline: largest/second listed value.
The \sys{} row uses settings adapted to standard LIBERO;
LIBERO-Pro and post-selection results use the frozen final system.
Configuration records: Appendix~\ref{app:eval}.}
\end{table}

\paragraph{LIBERO-Plus.} Table~\ref{tab:plus} gives the full breakdown. Camera
viewpoint ($66.5\%$) and robot initial state ($76.1\%$) are the hardest
categories, as they are for OpenVLA-OFT and $\pi_0$, and both perturb quantities
that analytic stages read directly. The run
used a configuration prepared for LIBERO-Plus rather than the routes of
Table~\ref{tab:libero}. All $1{,}568$ failures exhausted their step budget and
carry the same failure-layer label, so the explanation of the category
differences given above is a hypothesis. Of the records lost to
infrastructure, $242$ were rerun and replaced, the last $47$ with a tick limit of
$2{,}400$ instead of $600$, and the sensor-noise category ran with relaxed
simulator RPC waits; step budgets were unchanged and no valid failure was rerun.

\begin{table}[t]
\centering
\caption{\textbf{Outcome of every episode in the analyzed round.}}
\label{tab:denominator}
\footnotesize
\setlength{\tabcolsep}{6pt}
\begin{tabular}{@{}lr@{}}
\toprule
Outcome & Episodes \\
\midrule
Success (benchmark verdict true)   & 541 \\
Policy owned the episode           & 166 \\
Infrastructure (socket dropped)    & 88 \\
Plan refused                       & 4 \\
Other harness failure              & 1 \\
\midrule
Total, the reported denominator    & 800 \\
\midrule
Success over all episodes          & $67.6\%$ \,[64.3, 70.8] \\
Success excluding infrastructure   & $76.0\%$ ($712$ episodes) \,[72.7, 79.0] \\
\bottomrule
\end{tabular}
\end{table}

\begin{table}[t]
\centering
\caption{\textbf{Round histories} of Zetta (top;~\citealp{zetta2026}) and \sys{} on seeds $21$--$40$.}
\label{tab:rounds}
\footnotesize
\setlength{\tabcolsep}{5pt}
\begin{tabular}{@{}lrrr@{}}
\toprule
Zetta round & Goal-T & Goal-S & Mean \\
\midrule
\pizero{} & 31.0 & 38.0 & 34.5 \\
1 & 67.5 & 39.5 & 53.5 \\
2 & 89.5 & 70.5 & 80.0 \\
3 & 92.0 & 83.0 & 87.5 \\
4 & 92.5 & 89.0 & 90.8 \\
\midrule
\sys{} round & \multicolumn{2}{r}{Episodes} & Success (\%) \\
\midrule
\code{stat23}$^{a}$          & \multicolumn{2}{r}{800} & 60.0 \\
\code{stat25}$^{a}$          & \multicolumn{2}{r}{800} & 60.4 \\
\code{stat28}                & \multicolumn{2}{r}{800} & 66.3 \\
\code{stat29}                & \multicolumn{2}{r}{800} & 66.3 \\
\code{stat30}                & \multicolumn{2}{r}{800} & 67.6 \\
\code{stat31}$^{\times}$     & \multicolumn{2}{r}{780} & 59.7 \\
\code{stat33}$^{a}$          & \multicolumn{2}{r}{797} & 67.8 \\
\code{stat34}$^{a}$          & \multicolumn{2}{r}{800} & 68.4 \\
\code{stat35}$^{a}$          & \multicolumn{2}{r}{800} & 70.1 \\
\code{stat39}                & \multicolumn{2}{r}{800} & 72.8 \\
\code{stat40}$^{\times}$     & \multicolumn{2}{r}{800} & 70.8 \\
reported agent$^{a}$          & \multicolumn{2}{r}{800} & 74.25 \\
reported agent, remeasured   & \multicolumn{2}{r}{800} & 74.1 \\
\bottomrule
\end{tabular}
\par\vspace{2pt}{\footnotesize $^{\times}$Not promoted. $^{a}$Aggregate only.}
\end{table}

\begin{table}[t]
\centering
\caption{\textbf{Configurations on unperturbed LIBERO} ($2{,}000$ episodes each). LIBERO-adapted builds and the frozen LIBERO-Pro champion use different settings; the champion includes $1064$ tick-limit failures.}
\label{tab:liberoconfigs}
\footnotesize
\setlength{\tabcolsep}{2.6pt}
\begin{tabular}{@{}lccccc@{}}
\toprule
Method & Spatial & Object & Goal & Long & Avg. \\
\midrule
\pizero{}~$\star$                            & 99.0 & 99.0 & 98.6 & 95.4 & 98.00 \\
LIBERO \code{stdlive8c}              & 99.0 & 99.2 & 96.4 & 93.6 & 97.05 \\
LIBERO \code{formal1}                & 98.4 & 99.6 & 97.4 & 94.8 & 97.55 \\
LIBERO \code{formal2}                & 98.8 & 99.6 & 97.0 & 96.8 & 98.05 \\
Frozen champion (\code{088ef2ea}) & 86.2 & 30.4 & 69.8 & 0.4 & 46.70 \\
\bottomrule
\end{tabular}
\end{table}

\begin{table}[ht]
\centering
\caption{\textbf{Arms on both seed blocks.} Blocks are separate experiments.
The final development rates correspond to $74.25\%$ overall.}
\label{tab:variants}
\scriptsize
\setlength{\tabcolsep}{4.2pt}
\begin{tabular}{@{}llccccc@{}}
\toprule
Method & Seeds & Goal-T & Goal-S & 10-T & 10-S & Avg. \\
\midrule
\pizero{}, frozen                   & $1$--$20$  & 20.5 & 20.5 & 21.0 & 6.5  & 17.1 \\
\sys{} without evolution            & $1$--$20$  & 10.5 & 20.0 & 20.0 & 5.0  & 13.9 \\
\pizero{}, frozen                   & $21$--$40$ & 19.5 & 17.5 & 20.0 & 8.0  & 16.2 \\
\sys{}, analyzed round              & $21$--$40$ & 75.0 & 69.5 & 73.0 & 53.0 & 67.6 \\
\sys{}, \code{stat39}               & $21$--$40$ & 74.5 & 76.0 & 76.0 & 64.5 & 72.8 \\
\sys{}, archived final              & $21$--$40$ & 75.0 & 81.0 & 76.0 & 65.0 & 74.25 \\
\midrule
\multicolumn{7}{@{}l}{\textit{First attempts of the earlier 400-episode PhyAgentOS run}}\\
PhyAgentOS, GPT-4o-mini, first attempt & $1$--$10$ & 19.0 & 36.0 & 17.0 & 8.0 & 20.0 \\
PhyAgentOS, Qwen3-VL-4B, first attempt & $1$--$10$ & 21.0 & 34.0 & 18.0 & 8.0 & 20.3 \\
\bottomrule
\end{tabular}
\end{table}

\begin{table}[t]
\centering
\caption{\textbf{\sys{} on LIBERO-Plus}, one trial per task, with Wilson $95\%$ intervals.}
\label{tab:plus}
\footnotesize
\setlength{\tabcolsep}{4pt}
\begin{tabular}{@{}lrl@{}}
\toprule
Perturbation category & Successes & Success (\%) \\
\midrule
Camera viewpoints      & 1063/1599  & 66.5 \,[64.1, 68.8] \\
Robot initial states   & 1180/1550  & 76.1 \,[73.9, 78.2] \\
Objects layout         & 1322/1525  & 86.7 \,[84.9, 88.3] \\
Sensor noise           & 1391/1601  & 86.9 \,[85.1, 88.4] \\
Language instructions  & 1360/1537  & 88.5 \,[86.8, 90.0] \\
Light conditions       & 1098/1142  & 96.1 \,[94.9, 97.1] \\
Background textures    & 1048/1076  & 97.4 \,[96.3, 98.2] \\
\midrule
All tasks              & 8462/10030 & 84.4 \,[83.6, 85.1] \\
Mean over categories   &            & 85.5 \\
\midrule
Spatial tasks          & 2129/2402  & 88.6 \\
Object tasks           & 2237/2518  & 88.8 \\
Goal tasks             & 2097/2591  & 80.9 \\
LIBERO-10 tasks        & 1999/2519  & 79.4 \\
\bottomrule
\end{tabular}
\end{table}

\subsection{Post-Selection Evaluation}
\label{app:postselection}

\paragraph{Frozen systems and state banks.}
The final champion (\code{088ef2ea}), its prompt, library, parameters, budgets
and verifier were frozen before generating block C: $20$ new states for each
of $40$ task--perturbation cells ($800$ episodes). Block B comprises $261$
remaining never-inspected official states in $31$ cells; it is not a complete
official benchmark block. Both evaluations take place after system selection.
The four snapshots were chosen before examining these results and run with
their archived configurations. Their orchestration tick limits are $1200/2100$
for historical snapshots and $600$ for the champion; environment-step budgets
are unchanged. Reported development rates are $60.00$, $66.25$, $72.75$ and
$74.25\%$ for stat23, stat28, stat39 and the champion, respectively.

\begin{table}[ht]
\centering
\caption{\textbf{Post-selection success rates (\%) by suite.} C: $200$ newly
sampled states per suite. B: untouched official states, with $73$, $66$, $77$
and $45$ episodes in the listed suites, respectively.}
\label{tab:postselection_suites}
\small
\setlength{\tabcolsep}{6pt}
\begin{tabular}{@{}lrrrr@{}}
\toprule
Suite & C: \sys{} & C: \pizero{} & B: \sys{} & B: \pizero{} \\
\midrule
Goal-T & 79.0 & 26.5 & 78.1 & 26.0 \\
Goal-S & 78.0 & 19.5 & 81.8 & 25.8 \\
10-T & 77.5 & 17.5 & 75.3 & 7.8 \\
10-S & 66.5 & 6.5 & 71.1 & 2.2 \\
\midrule
Episodes & 800 & 800 & 261 & 261 \\
Success (\%) & 75.2 & 17.5 & 77.0 & 16.5 \\
\bottomrule
\end{tabular}
\end{table}

\paragraph{Paired results and snapshot transfer.}
New-state gains over the frozen policy are $52.5$, $58.5$, $60.0$ and $60.0$
percentage points across Goal-T, Goal-S, 10-T and 10-S. All four frozen snapshots
retain their development ordering (Spearman $\rho=1.00$). From stat23 to the
champion, the improvement is $14.25$ points in development and $14.375$ on new
states. The last increment, stat39 to champion, adds $8$ successes over $800$ episodes with
$12$ versus $4$ discordant wins ($p=0.0768$), which does not establish a positive
last-step effect at the $0.05$ level. The cumulative gain and the final increment
are separate claims. Table~\ref{tab:postselection_pairs} reports paired differences
with cell-bootstrap intervals over task--perturbation cells.

\begin{table}[ht]
\centering
\caption{\textbf{Paired post-selection comparisons.} W/L are exclusive wins
of the first/second system. Intervals are cell-bootstrap $95\%$ intervals;
$p$ is a two-sided exact test of discordant pairs.}
\label{tab:postselection_pairs}
\scriptsize
\setlength{\tabcolsep}{5pt}
\begin{tabular}{@{}llrrrrl@{}}
\toprule
Block & Comparison & $n$ & W/L & $\Delta$ (pp) & $95\%$ CI & $p$ \\
\midrule
C & Champion vs. \pizero{} & 800 & 473/11 & 57.75 & [45.75, 70.00] & $3.13\times10^{-124}$ \\
B & Champion vs. \pizero{} & 261 & 161/3 & 60.54 & [44.44, 75.27] & $6.29\times10^{-44}$ \\
C & Champion vs. stat23 & 800 & 137/22 & 14.38 & [6.62, 23.62] & $1.70\times10^{-21}$ \\
C & Champion vs. stat28 & 800 & 93/18 & 9.38 & [3.88, 16.00] & $2.26\times10^{-13}$ \\
C & Champion vs. stat39 & 800 & 12/4 & 1.00 & [0.00, 2.62] & 0.0768 \\
C & stat28 vs. stat23 & 800 & 65/25 & 5.00 & [0.12, 11.50] & $2.97\times10^{-5}$ \\
C & stat39 vs. stat28 & 800 & 86/19 & 8.38 & [3.50, 14.37] & $2.30\times10^{-11}$ \\
\bottomrule
\end{tabular}
\end{table}

\paragraph{Execution accounting.}
Block C uses post-protocol amendment A1. The affected host, Pro4, had observed
load averages of $77$--$101$; all $187$ flagged champion episodes originated
there. The replacement rule was host membership: rerun every C cell--system job
assigned to Pro4, including successful episodes, rather than selecting failures.
All $54$ jobs ($19$ champion and $35$ historical-snapshot jobs) were rerun on
U2002, Pro1 or Pro3 with at most $12$ concurrent slots. Each replacement refers
to the same system and state bank; other hosts' results are retained. The original
and A1 records are both preserved. The mapping below gives successes and flagged
episodes out of $800$ per arm.

\begin{center}
\small
\textbf{A1 replacement accounting ($800$ episodes per arm).}\\[2pt]
\begin{tabular}{@{}lrr@{}}
\toprule
System & Original: successes / flags & A1: successes / flags \\
\midrule
stat23 & 487 / 18 & 487 / 9 \\
stat28 & 523 / 5 & 527 / 3 \\
stat39 & 574 / 22 & 594 / 2 \\
Champion & 472 / 187 & 602 / 2 \\
\bottomrule
\end{tabular}
\end{center}

The frozen-policy and five-capability-withheld arms are unchanged by A1
($17.5\%$ and $65.8\%$, with $0$ and $51$ flags). All retained flags count as
failures; no episode is removed from a denominator. Block B is unchanged
(champion: $77.0\%$, $2$ flags; frozen policy: $16.5\%$, no flags).

\paragraph{Five-capability transfer check.}
Withholding the five evolved capabilities on block C gives $65.8\%$,
versus $75.2\%$ for the champion: $85$ champion-only and $9$ treatment-only
wins ($9.5$ points, $p=1.2\times10^{-16}$). The treatment has $51$ flagged
failures versus $2$ for the champion, so this contrast includes runtime effects
of the removal. It is separate from the seven-contact-skill ablation on the
development block in Appendix~\ref{app:mechanism}.

\section{Additional Results}
\label{app:results}

\begin{figure}[ht]
\centering
\includegraphics[width=\textwidth]{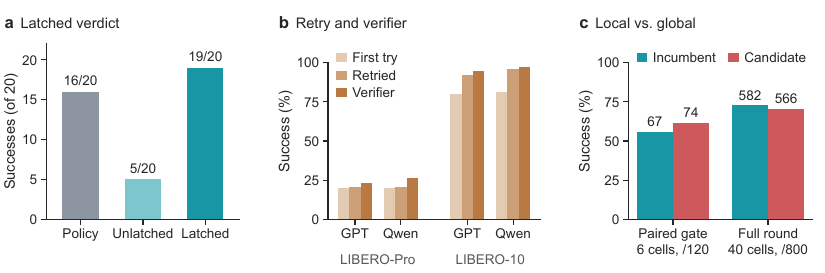}
\vspace{-5mm}
\caption{\textbf{Supporting panels.} \textbf{a}~Replayed stove requests.
\textbf{b}~PhyAgentOS retries and verifier. \textbf{c}~Paired gate against full
round for candidate C2.}
\label{fig:suppl}
\end{figure}

This section gives the full account behind the insights of
Sections~\ref{sec:fastslow} to~\ref{sec:causal}, including the
records that the main text summarizes and the diagnostic comparisons moved here
(Fig.~\ref{fig:suppl}).

\begin{figure}[ht]
\centering
\includegraphics[width=0.375\textwidth]{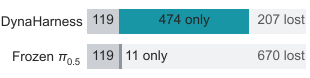}
\vspace{-2mm}
\caption{\textbf{Paired outcomes} with the frozen \pizero{} on the same $800$
episodes (second measurement, $74.1\%$); each bar: won by both, won by this
arm only, lost.}
\label{fig:paired}
\end{figure}

\subsection{Capability and Closed-Loop Execution Ablations}
\label{app:mechanism}

\paragraph{Protocol and interventions.}
Each arm covers the same $800$ development episodes (seeds $21$--$40$), paired
by task, perturbation and initial state. The capability-removal arms share frozen
\pizero{}, Qwen3-VL-4B, prompts, parameters and environment-step budgets with
their concurrent control. The same-skill A2static and A2seq comparisons reuse
that $74.0\%$ control (A2ctrl), separately from the $74.25\%$ development
headline and post-selection evaluations. A2static retains the original one-step
planner interface with nominal replanning; A2seq freezes a full initial sequence,
as detailed below.
Seven analytic contact skills are removed from the
registered action set in the analytic arm; six recovery/intervention capabilities
are removed in the recovery arm. Removing both retains the planner, grounding,
completion checks, latching, budgets and frozen VLA. Capability removal is
verified from executed skill traces. Contrasts are not additive decompositions.

\begin{table}[ht]
\centering
\caption{\textbf{Capability and executor ablations by suite.} All arms have $800$ episodes,
$200$ per suite. Flags denote tick-limit or transport contamination, counted as
failures; a dash denotes an unreported count. The bare-policy row is the archived
same-block reference.}
\label{tab:mechanism_counts}
\scriptsize
\setlength{\tabcolsep}{5pt}
\begin{tabular}{@{}lrrrrrr@{}}
\toprule
Arm & Goal-T & Goal-S & 10-T & 10-S & Total & Flags \\
\midrule
Full \sys{} & 150 & 161 & 152 & 129 & 592 & 7 \\
\rowcolor{hmg!10}Nominal replanning (A2static) & 130 & 150 & 117 & 114 & 511 & --- \\
Frozen sequence (A2seq) & 130 & 159 & 115 & 106 & 510 & 0 \\
No analytic contact skills & 54 & 32 & 31 & 16 & 133 & 8 \\
No recovery/intervention & 149 & 161 & 145 & 130 & 585 & 7 \\
Neither capability group & 47 & 31 & 36 & 12 & 126 & 4 \\
VLA unavailable & 130 & 159 & 151 & 124 & 564 & 221 \\
Inert-switch replicate (A2e1) & 150 & 162 & 152 & 129 & 593 & 7 \\
Bare \pizero{} & 39 & 35 & 40 & 16 & 130 & --- \\
\bottomrule
\end{tabular}
\end{table}

\begin{table}[ht]
\centering
\caption{\textbf{Paired executor and capability comparisons.} W/L are wins
exclusive to the first/second arm; differences are first minus second. A uses
cell-bootstrap $95\%$ intervals. Full/A2seq bootstrap estimates use $20{,}000$
whole-cell resamples, seed $20260926$. All $p$ values are exact two-sided sign tests.}
\label{tab:mechanism_pairs}
\scriptsize
\setlength{\tabcolsep}{5pt}
\textbf{A. Aggregate contrasts}\par
\begin{tabular}{@{}lrrrr@{}}
\toprule
Comparison (Full minus treatment unless named) & W/L & $\Delta$ (pp) & $95\%$ CI & Exact $p$ \\
\midrule
\rowcolor{hmg!10}Nominal replanning (A2static) & 89/8 & 10.125 & [3.375, 18.250] & $2.00\times10^{-18}$ \\
Frozen sequence (A2seq) & 93/11 & 10.25 & [3.50, 18.25] & $2.49\times10^{-17}$ \\
A2static minus A2seq & 11/10 & 0.125 & [$-3.000$, 2.875] & 1.00 \\
No analytic contact skills & 469/10 & 57.38 & [45.62, 69.00] & $2.09\times10^{-124}$ \\
No recovery/intervention & 15/8 & 0.88 & [$-0.38$, 2.62] & 0.21 \\
Neither capability group & 473/7 & 58.25 & [46.62, 69.62] & $7.25\times10^{-130}$ \\
VLA unavailable & 32/4 & 3.50 & [0.00, 9.25] & $1.94\times10^{-6}$ \\
Inert-switch replicate (A2e1) & 5/6 & $-0.125$ & --- & 1.00 \\
\bottomrule
\end{tabular}
\par\smallskip
\textbf{B. Full versus A2static: paired outcomes}\par
\begin{tabular}{@{}lrrrr@{}}
\toprule
Suite & W/L & Ties & $\Delta$ (pp) & Exact $p$ \\
\midrule
All & 89/8 & 703 & 10.125 & $2.00\times10^{-18}$ \\
Goal-T & 20/0 & 180 & 10.0 & $1.91\times10^{-6}$ \\
Goal-S & 14/3 & 183 & 5.5 & 0.0127258 \\
10-T & 35/0 & 165 & 17.5 & $5.82\times10^{-11}$ \\
10-S & 20/5 & 175 & 7.5 & 0.00407732 \\
\bottomrule
\end{tabular}
\par\smallskip
{\scriptsize Paired Wald $95\%$ CIs: Full/A2static $[7.816,12.434]$ pp;
A2static/A2seq $[-0.998,1.248]$ pp.}\par\smallskip
\textbf{C. Full versus A2seq: aggregate and suite-level estimates}\par
\setlength{\tabcolsep}{3pt}
\begin{tabular}{@{}lrrrrr@{}}
\toprule
Suite & W/L & $\Delta$ (pp) & Wald $95\%$ CI & Cell-bootstrap $95\%$ CI & Exact $p$ \\
\midrule
All & 93/11 & 10.25 & [7.85, 12.65] & [3.50, 18.25] & $2.49\times10^{-17}$ \\
Goal-T & 20/0 & 10.00 & [5.84, 14.16] & [0.00, 29.00] & $1.91\times10^{-6}$ \\
Goal-S & 9/7 & 1.00 & [$-2.92$, 4.92] & [$-2.50$, 5.00] & 0.803619 \\
10-T & 37/0 & 18.50 & [13.12, 23.88] & [2.00, 40.01] & $1.46\times10^{-11}$ \\
10-S & 27/4 & 11.50 & [6.28, 16.72] & [1.00, 24.00] & $3.40\times10^{-5}$ \\
\bottomrule
\end{tabular}
\end{table}

\paragraph{Where performance changes.}
Analytic-skill removal costs $96$, $129$, $121$ and $113$ successes across the
four suites, reducing every suite to $27.0\%$ success or below. Recovery removal
costs $7$ successes overall; $7$ are in 10-T, with the other suites changing by
$-1$, $0$ and $+1$. Planning and the execution contract alone, with both groups
removed, score $15.8\%$, near the bare policy's $16.2\%$.
With the library retained, A2static and A2seq lose $81$ and $82$ net successes
to Full. A2static adds only one net success to A2seq, so nominal replanning
alone does not reproduce the dynamic executor's performance. The $74.0\%$ versus $74.1\%$ inert-switch replicate ($800$ episodes per arm,
$5/6$ discordant pairs, $p=1$) provides a same-program variation reference;
its difference is not subtracted from these effects.

\paragraph{Capability provenance.}
The initial library contained policy execution, pick-and-place and perception.
The dated registry records later additions for pushing (September 5), drawer
sliding and knob turning (September 6), keyframe recovery (September 7), and door
swinging (September 9). Existing execution mechanisms were also revised, including
completion latching and drawer grasp geometry. Thus the seven-contact-skill
removal tests the final library, which combines initial and subsequently added
or refined capabilities. The separate five-capability removal tests documented
additions; neither intervention removes the entire development process.

\paragraph{Policy invocation.}
In $770$ archived control event stores, $570$ episodes do not invoke \pizero{}:
$96.7\%$ succeed, versus $6.0\%$ among the $200$ episodes that do.
The original $800$-episode development event set similarly has $73\%$ zero-policy
episodes. The new-state event stores cover $777/800$ episodes, of which $570$
($73.4\%$) never invoke the policy; the official-state block has $193/261$
($73.9\%$). These are conditional descriptions: policy calls concentrate in
hard episodes after analytic execution struggles, rather than forming a randomized
comparison of policy use. This execution pattern is consistent with the role of
analytic capabilities as the main source of task competence and the VLA as an
available learned route. The same-skill executor comparisons separately evaluate
closed-loop use of the library (Appendix~\ref{app:static}).

\begin{table}[ht]
\centering
\caption{\textbf{Routing, coverage and execution diagnostics.} A: prior routing
archive. B: result and trace coverage for the executor comparisons. C: common
$783$ Full/A2static pairs. D: all $800$ A2static/A2seq pairs. Event counts are
descriptive, may overlap, and do not identify individual causal contributions.}
\label{tab:routing_new}
\scriptsize
\setlength{\tabcolsep}{5pt}
\begin{tabular}{@{}lrrrr@{}}
\toprule
\multicolumn{5}{@{}l}{\textbf{A. Prior routing archive}}\\
Arm/block & Event stores & Zero VLA (\%) & VLA calls/ep. & Planner calls/ep. \\
\midrule
A2 full control & 770 & 74 & 2.39 & 3.68 \\
No analytic contact skills & 800 & 0 & 18.34 & 4.12 \\
No recovery/intervention & 750 & 75 & 2.11 & 4.09 \\
Neither capability group & 800 & 0 & 18.52 & 4.06 \\
VLA unavailable & 705 & $\approx100$ & 0.00 & 17.01 \\
Champion, B & 261 & 74 & 2.30 & 3.02 \\
Champion, C & 777 & 73 & 2.19 & 3.84 \\
\bottomrule
\end{tabular}
\par\smallskip
\begin{tabular}{@{}lrrr@{}}
\toprule
\multicolumn{4}{@{}l}{\textbf{B. Same-skill executor coverage}}\\
Arm & Valid result rows & Complete event traces & Paired traces \\
\midrule
Full (A2ctrl) & 800 & 783 & 783 \\
A2static & 800 & 800 & 783 \\
A2seq & 800 & 800 & 783 \\
\bottomrule
\end{tabular}
\par\smallskip
\begin{tabular}{@{}lrr@{}}
\toprule
\multicolumn{3}{@{}l}{\textbf{C. Full/A2static: $783$ common traces}}\\
Metric & Full & A2static \\
\midrule
Planner calls & 2{,}860 & 1{,}829 \\
Post-initial planner calls & 1{,}747 & 1{,}046 \\
Failure/escalation replans & 566 & 0 \\
Capability substitutions & 471 & 0 \\
Capability-ID switches after substitution & 327 & 0 \\
Recovery insertions & 141 & 0 \\
Verifier-stagnation branch changes & 352 & 0 \\
Fixed-step retries & 0 & 534 \\
Fixed-plan reexecutions & 0 & 255 \\
Fixed retry exhaustion & 0 & 241 \\
Same-capability/arguments retries after failure & 26 & 587 \\
Rejected plans & 205 & 59 \\
Environment steps & 243{,}530 & 190{,}107 \\
\midrule
\multicolumn{3}{@{}l}{\textbf{D. A2static/A2seq: all $800$ traces}}\\
Metric & A2static & A2seq \\
\midrule
Planner calls & 1{,}846 & 800 \\
Post-initial planner calls & 1{,}046 & 0 \\
Capability substitutions & 0 & 0 \\
Failure/escalation replans & 0 & 0 \\
Recovery insertions & 0 & 0 \\
Verifier-stagnation branch changes & 0 & 0 \\
\bottomrule
\end{tabular}
\end{table}

\paragraph{VLA-unavailable condition.}
Removing the sole policy-calling capability gives $70.5\%$.
Its $3.5$-point loss includes routing and planning effects: the planner makes
$17.01$ calls and $4.69$ proposals naming a removed capability per archived
episode, with $221$ flagged failures. The loss is therefore only an upper-bound
proxy for VLA contribution, combining policy availability with the routing and
runtime effects of removal.

\paragraph{Intervention at matched failure states.}
A replay study compared an alternative action with the control suffix from the
same state: $22$ alternative-only wins versus $14$ control-only wins
($p=0.24$), or $13.0\%$ versus $9.4\%$ success. A subsequent $120$-episode,
$342$-branch validation on seeds $41$--$70$ found $22.2\%$ success over $54$
episodes for both \code{close\_and\_lift} and its control, and $7.1\%$ versus
$19.0\%$ over $42$ episodes for \code{release\_and\_retreat}. These measurements do not show that a single
substitution reliably recovers the observed failure states.

\paragraph{Intervention bookkeeping.}
The analytic treatment uses registry-removal revision N2, verified from executed
skill traces. Grounding, command budgets and leases remain enabled in all treatments. The recovery arm
also retains planning and verification, so it is not a separate state-machine
baseline. Attribution and admission govern development and are not isolated by
these frozen-runtime interventions.

\paragraph{Nominal one-step replanning.}\label{app:static}
A2static retains the analytic capability library and Full's original one-step
planner interface. After successful completion of a nominal skill, it may query
the planner using the updated observation for the next nominal skill. It disables
failure/escalation-triggered replanning, capability substitution, dynamic
reordering, recovery insertion and verifier-triggered branch changes. Failures
instead invoke fixed retries and plan reexecution. Thus the three-way comparison
progresses from frozen sequencing (A2seq), through nominal replanning (A2static),
to Full's state-dependent dynamic execution, without removing analytic skills.

Full and A2static share $503$ successes and $200$ failures, with $89$ Full-only
and $8$ A2static-only successes. A2static and A2seq share $500$ successes and
$279$ failures, with $11$ A2static-only and $10$ A2seq-only successes.
Table~\ref{tab:mechanism_pairs} reports aggregate paired uncertainty and
suite-level outcomes. The $0.125$-point nominal-replanning increment has Wald
$95\%$ CI $[-0.998,1.248]$ and cell-bootstrap CI $[-3.000,2.875]$; this is
an observed small difference, not an equivalence test. Full's $10.125$-point
gain over A2static evaluates its dynamic mechanisms jointly and is broader than
the $0.9$-point effect of withholding dedicated recovery/intervention capabilities.
These are comparisons on the development block, not unseen-task evaluations.

All three arms have $800$ outcomes on the same official initial states.
A2static and A2seq have $800$ event traces each; paired diagnostics with Full
use its $783$ available traces. Table~\ref{tab:routing_new} distinguishes this
common subset from the complete A2static/A2seq comparison. The latter records
$1{,}046$ additional planner calls in A2static, while substitutions, failure
replans, recovery insertions and verifier-stagnation branches remain zero in
both arms. Nominal online planning therefore occurs as intended, but does not
reproduce Full's aggregate performance. The event counts document execution
behavior rather than assigning success gains to individual branches.

\paragraph{Same-skill sequential executor.}\label{app:seq}
The A2seq protocol dated September 26, 2026 evaluates $40$ LIBERO-Pro cells
on official initial states indexed by seeds $21$--$40$, paired with the
existing A2ctrl measurement. All analytic capabilities, recovery skills, tools
and frozen \code{pi05\_libero} remain mounted. Both use Qwen3-VL-4B, the same
initial observation/state contents, grounding, skill-completion predicates,
benchmark verifier, task-success criterion and low-level safety gates.
Shared planner settings are temperature $0.1$, $2048$ tokens per call,
$90$\,s timeout, $180$\,s plan validity and one serialization-repair allowance.
Environment budgets remain $300/520$ steps and $600$ ticks.

A2seq makes one initial planning request, freezes a complete ordered sequence
with explicit arguments, and executes it strictly in order. A failed step is
retried once; after exhausting that retry, the same single-step plan is replayed
once with one further same-step retry. Precondition blockage for $12$ decision
ticks counts as a failure, and an unresolvable binding terminates execution.
Post-execution planner calls, online replanning, reordering, recovery insertion,
capability substitution and unplanned dispatch are disabled. Recovery capabilities
can still appear in the initial plan. The next-capability output schema of Full
is replaced by a complete-sequence schema; the contrast therefore includes
upfront planning and the loss of receding-horizon planning, as well as execution
policy. It does not isolate one scheduler component.

\paragraph{Paired outcome and scope.}
Over $800$ episodes per arm, Full scores $74.00\%$ (Wilson $95\%$ CI
$[70.85,76.92]$), versus $63.75\%$ for A2seq ($[60.36,67.01]$). The $93$ Full-only and
$11$ A2seq-only successes yield $+10.25$ points; $499$ pairs both succeed and
$197$ both fail. Table~\ref{tab:mechanism_pairs} reports paired uncertainty;
this supporting comparison pools all $800$ pairs, with suite-level estimates
describing its distribution. A2seq ran on U2002, whereas A2ctrl is a historical
multi-host measurement. Date, machine and unseeded planner or policy stochasticity
remain nuisance factors. This development-block comparison evaluates execution
design, not blind generalization to unseen tasks.

\paragraph{Completeness and execution audit.}
Both arms have $800$ valid measurement rows, with zero initial-state pairing
mismatches and no missing results. Full's $7$ tick-limit stops count as failures;
A2seq has none. Full has complete trajectories for $783$ episodes; the $17$
missing traces are \code{libero\_goal\_task[7]}, seeds $24$--$40$. Their valid
results remain in the $800$-pair success statistics. A2seq has all $800$ traces;
event comparisons use only the $783$ common pairs (Table~\ref{tab:routing_new}).
Before scoring, $3$ paired smoke cases, $187$ unit/regression tests and $58$
smoke/consistency checks passed; $297$ source/configuration files were frozen
and verified, with matching BDDL and official-state hashes across hosts.
The post-run audit verified one initial planner call per A2seq episode, frozen
ordering and arguments, bounded retries, and no dynamic branch or unplanned
dispatch. Historical champion results were not modified.

\begin{table}[!htb]
\centering
\caption{\textbf{Full versus A2seq: execution diagnostics.} A: directly recorded
whole-run totals. B: events or steps on the $783$ pairs with both trajectories.
Rows may overlap and are not independent causal effects.}
\label{tab:seq_events}
\footnotesize
\setlength{\tabcolsep}{6pt}
\begin{tabular}{@{}lrr@{}}
\toprule
Metric & Full & A2seq \\
\midrule
\multicolumn{3}{@{}l}{\textbf{A. Whole run: $800$ episodes per arm}}\\
Planner calls & 2{,}875 & 800 \\
Environment steps & 244{,}523 & 188{,}038 \\
Environment-budget exhaustion & 201 & 17 \\
Tick-budget exhaustion & 7 & 0 \\
\midrule
\multicolumn{3}{@{}l}{\textbf{B. Common event traces: $783$ pairs}}\\
Capability substitutions & 471 & 0 \\
Dispatched capability-ID switches after substitution & 327 & 0 \\
Same-ID substitutions & 70 & 0 \\
Failure/escalation replans & 566 & 0 \\
Fixed-plan reexecutions & 0 & 230 \\
Fixed retry exhaustion & 0 & 218 \\
Fixed step retries & 0 & 494 \\
Post-initial planner calls & 1{,}747 & 0 \\
Recovery insertions & 141 & 0 \\
\quad Planned & 93 & 0 \\
\quad Scheduler & 48 & 0 \\
Verifier-stagnation branch changes & 352 & 0 \\
Rejected plans & 205 & 36 \\
Environment steps & 243{,}530 & 188{,}038 \\
Environment-budget exhaustion & 200 & 17 \\
\bottomrule
\end{tabular}
\end{table}

\paragraph{Execution differences.}
Table~\ref{tab:seq_events} confirms that Full uses replanning, substitution and
recovery insertion, while A2seq uses fixed retries and plan replay. The counts
verify the intended execution-policy difference; success is assessed by the
paired outcomes, rather than by treating event types as independent contributions.

\begin{table}[!htb]
\centering
\caption{\textbf{First dispatch divergence on $783$ trace-matched pairs.}
F/S: Full-only/A2seq-only success; both columns denote shared outcomes.
Categories are descriptive associations with the eventual result.}
\label{tab:seq_divergence}
\footnotesize
\setlength{\tabcolsep}{5pt}
\begin{tabular}{@{}lrrrrr@{}}
\toprule
First divergence & Pairs & F & S & Both succeed & Both fail \\
\midrule
Identical dispatches & 468 & 0 & 0 & 468 & 0 \\
Spent-capability substitution & 109 & 16 & 3 & 12 & 78 \\
Planned recovery insertion & 84 & 18 & 1 & 0 & 65 \\
Initial planner choice differs & 61 & 22 & 6 & 19 & 14 \\
Scheduler recovery insertion & 27 & 2 & 0 & 0 & 25 \\
Online planner selection & 19 & 10 & 1 & 0 & 8 \\
Failure/escalation replan & 7 & 5 & 0 & 0 & 2 \\
Nominal fallback & 4 & 0 & 0 & 0 & 4 \\
Held-object mismatch substitution & 3 & 3 & 0 & 0 & 0 \\
Precondition substitution & 1 & 1 & 0 & 0 & 0 \\
\midrule
Total with both traces & 783 & 77 & 11 & 499 & 196 \\
\bottomrule
\end{tabular}
\end{table}

Across the $468$ pairs with identical dispatches, both executors succeed.
Differences concentrate on the remaining pairs, where Full changes capabilities,
inserts recovery or makes additional planning decisions; $61$ pairs already
differ in their initial planner choice (Table~\ref{tab:seq_divergence}).
This supports the aggregate closed-loop comparison while retaining the planning
format distinction. The $17$ pairs lacking Full traces have no assigned category.

\subsection{Fast-Slow Robot Agent}

\begin{figure}[ht]
\centering
\includegraphics[width=0.4\textwidth]{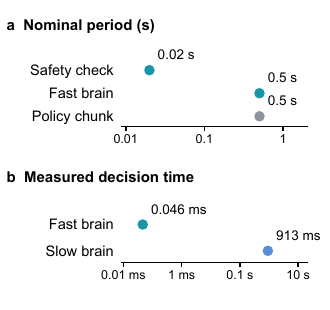}
\caption{\textbf{Execution schedule and decision latency.} \textbf{a}~Nominal
execution periods. \textbf{b}~Median scheduler decision and planner-call latencies
in the $800$-episode run (Table~\ref{tab:latency}).}
\label{fig:timescales}
\end{figure}

\paragraph{The stove window.}
In \emph{turn off the stove}, the policy turned the knob off at step $44$ and
back on by step $55$. Across $20$ replays, sampling every $20$ steps without
latching succeeds on $25\%$; the bare policy achieves $80\%$, and latching with
a chunk-level read achieves $95\%$ (Fig.~\ref{fig:suppl}a). This replay changes both sampling and
latching. The paired ablation in Appendix~\ref{app:ablation} holds the sampling period
fixed. A separate grounding example records $35$ to $39$ refusals when
the planner named a book that neither camera observed.

\paragraph{Decision granularity.}
The fast brain operates on a nominal $0.5$\,s control schedule, while the slow
brain is called on demand. The Harness VLA diagnostic runs in
Table~\ref{tab:hvla} took $31$ to $37$ turns and $348$ to $505$\,s per episode,
including tool execution and simulation. These wall-clock measurements describe
runtime overhead; temporal coverage of physical events depends on the
observations and environment steps within each tool call.

\paragraph{Measured rates and wall clock.} Table~\ref{tab:latency} gives the
latency of each decision and the wall clock of an episode, read from the event
store of the second measurement of the reported agent ($74.1\%$). The runs used $22$ channels on four GPUs, so these
are latencies under load. The $0.046$\,ms fast-brain median measures the
scheduler decision routine (\code{scheduler.decision.latency\_ms}); the
$913$\,ms slow-brain median measures a planner call
(\code{slow\_brain.completed.latency\_ms}). Figure~\ref{fig:teaser}c and
Fig.~\ref{fig:timescales} use these same values. Decision computation, the
nominal $0.5$\,s schedule and end-to-end tool execution are distinct quantities.
Tool durations include
simulator stepping and rendering: \code{vla\_act} takes $1{,}670$\,ms for a
$20$-step call (two chunks of ten actions), \code{guarded\_descend} $2{,}030$\,ms for $34$ steps and
\code{turn\_knob} $11{,}980$\,ms for $174$ steps, while \code{verify\_outcome}
takes $12$\,ms. The forward latency of \pizero{} alone is not recorded, so it
is not separable from simulation here.

In a separate diagnostic run, we measured inter-turn latency for Harness VLA driven by Codex (\code{gpt-5.5}, reasoning
effort \code{none}) on one \code{libero\_object\_swap} episode: over $20$
inter-turn decisions taken from two identical runs, the time from a completed
execution-time tool call to the start of the next was $4.20$\,s at the median,
$4.66$\,s on average and $7.03$\,s at the 95th percentile, cross-checked against
the agent's own event clock ($4.33$\,s median). The measurement excludes the
previous tool's execution, cold start, failed calls and batched calls, so it is
the agent's own decision time. It is our reproduction and not a number that paper
reports. The task, scaffold and workload differ from the \sys{} latency run.
The $6.683$\,s value in Fig.~\ref{fig:teaser}c instead comes from replacing
\sys{}'s slow brain with Sonnet 5 through Claude Code: $107$ calls in the
$20$-episode pilot (Table~\ref{tab:upper}), whose paired Qwen median is
$1.124$\,s. These are separate runs and measurement scopes, not a matched
cross-system speed comparison.

\begin{table}[ht]
\centering
\caption{\textbf{Latency and wall clock of the reported agent}, $22$ channels running concurrently.}
\label{tab:latency}
\footnotesize
\setlength{\tabcolsep}{5pt}
\begin{tabular}{@{}lrrrrr@{}}
\toprule
Component & Calls & p50 & p90 & p99 & max \\
\midrule
Slow brain (Qwen3-VL-4B) & 3{,}090 & 913\,ms & 1{,}277\,ms & 1{,}865\,ms & 5{,}581\,ms \\
Fast brain   & 47{,}199 & 0.046\,ms & 0.056\,ms & 0.071\,ms & 5.58\,ms \\
Fast brain, learned artifact & -- & 0.49\,ms & -- & -- & -- \\
\midrule
Episode wall clock & Mean & p50 & p90 & Success & Failure \\
\midrule
Goal-T & 19.2\,s & 15.5\,s & 36.0\,s & 13.7\,s & 35.4\,s \\
Goal-S & 22.1\,s & 18.5\,s & 35.0\,s & 18.8\,s & 36.6\,s \\
10-T   & 37.6\,s & 35.0\,s & 54.1\,s & 30.8\,s & 59.1\,s \\
10-S   & 46.2\,s & 38.0\,s & 58.5\,s & 33.7\,s & 69.5\,s \\
All & 31.3\,s & 29.8\,s & -- & -- & -- \\
\bottomrule
\end{tabular}
\end{table}

\paragraph{Learning the fast brain.}\label{app:learned} The evidence stores were designed to make the fast brain learnable, and a first
attempt shows what the current data supports. Pooled and de-duplicated, the
stores hold $2.59$ million decisions over $27{,}581$ episodes, each with the state
the fast brain saw, the capability mask, the action and its reason. Because the
fast brain is deterministic, $99.2\%$ of these decisions continue the current
command and the rest come from a single rule, so the logs offer almost no
counterfactual actions. Advantage-weighted regression therefore reproduces the
rule, agreeing with it on $99.93\%$ of held-out decisions. Its one systematic
departure is informative: in $10.7\%$ of the states where the rule intervenes,
the learned policy prefers to continue, and these states lie late in the budget
after about two failed interventions, with a mean return of $-0.05$ against
$+0.33$ elsewhere. Objectives that move further from the logged action place
probability on actions absent from the corpus and agree with the rule on $87.5\%$
of held-out decisions. To broaden action coverage, an $\epsilon$-greedy pass has collected $279$ such decisions in $160$
episodes, which are excluded from every success rate here. The paired online
comparison has since run to completion: over the full $800$ episodes the learned
and the fast brain both score $593$ ($p = 1.0$), and on the cells the
learned artifact can reach it escalates less often ($12.2$ against $15.4$ times
per episode). Two diagnostics characterize the training signal. The reward's progress term is
counted only when the progress signal is trusted, which holds for $9.1\%$ of
transitions, and the value changes between adjacent decisions in $0.04\%$ of
them, so the term is non-zero in about $0.004\%$ of decisions and the scheduler
optimizes little more than a time penalty and the final outcome. A candidate
knob that raises that density also drives the stagnation test, whose trigger
rate rose from $1.3\%$ to $16.4\%$ in one cell, so it was not admitted. $91.0\%$ of the $47{,}199$ recorded decisions had more than one eligible
action, while the training corpus contained little action diversity. One limit is structural: whether a segment stays with an
analytic skill or goes to the VLA is a plan step, outside the scheduler's action
space.

\begin{table}[ht]
\centering
\caption{\textbf{Completion latching and five evolved capabilities.} Separate
paired campaigns on development seeds $21$--$40$, $800$ episodes per arm.}
\label{tab:ablation}
\small
\setlength{\tabcolsep}{6pt}
\begin{tabular}{@{}lrrr@{}}
\toprule
Variant & Success (\%) & Net episodes & Paired $p$ \\
\midrule
Latching control & 73.1 & --- & --- \\
Without latched verdict & 69.6 & $-28$ & $6.2\times10^{-5}$ \\
Five-capability control & 73.6 & --- & --- \\
Without five evolved capabilities & 64.8 & $-71$ & $<10^{-13}$ \\
\bottomrule
\end{tabular}
\end{table}

\paragraph{Paired ablations.}\label{app:ablation} The two earlier ablations summarized in
Section~\ref{sec:causal} run treatment and control in one campaign over the same
$800$ development episodes, paired on suite, task and seed, and are read with an
exact McNemar test on the discordant episodes. The control arms scored $585$ and
$589$ of $800$. They are separate runs and both sit below the $594$ of the
reported agent, so only the difference inside each campaign is interpreted and
no row of Table~\ref{tab:ablation} is compared with Table~\ref{tab:main}.
The latch ablation changes Eq.~\ref{eq:verify} alone, leaving the sampling
period, the budget and the lease untouched. The capability ablation removes five
entries from the library $\mathcal{L}$ with corresponding planner-catalog
filtering; the cells reachable by
at least one of them were listed before the run, and the observed losses are reported by this predefined split.

In the latch ablation, the benchmark predicate becomes true and later false again
in $131$ of the $800$ episodes, and such transient outcomes occur in $17$ of the
$40$ cells; those $17$ cells account for a net $-34$ successes while the other
$23$ together move by $+6$. In \code{10\_task[5]}, which has $19$ such episodes,
unlatching costs $15$ successes. Paired on the same episodes, $38$ are won only
with the latch and $10$ only without it. In the capability ablation, $82$
episodes are won only by the full agent and $11$ only without the five
capabilities; by suite the loss is Goal-S $-41$, 10-T $-20$, 10-S $-11$ and
Goal-T $+1$. Capability filtering visible to the planner was corrected before
this run, so the treatment no longer makes the planner propose what it cannot
call (next paragraph).

\paragraph{Five-capability treatment configuration.}\label{app:bb2}
The treatment filters removed capabilities from the planner catalog but leaves
other execution paths active. Its $64.75\%$ score includes $51$ flagged failure
episodes; the logs also contain $539$ disabled-skill rejection events from
non-planner paths. Rejection events are not episode counts, and their contribution
to the flagged failures is not isolated. The paired contrast therefore measures
the system-level effect of withholding the entries, including routing and runtime
consequences, rather than their physical execution benefit alone. Mechanism
changes remain enabled in both arms. The earlier unfiltered diagnostic
($61.375\%$) is excluded from this comparison.

\paragraph{Upper-model sensitivity.}\label{app:upper} We replaced the slow brain
with Claude Sonnet 5 through Claude Code, and froze everything else.
The arm is not a pure model swap: the agent brings its own scaffold of three
turns and about $74.6$k prompt tokens per call, against one turn and about $4$k
for Qwen3-VL-4B, and the output schema is carried in the prompt rather than
enforced by the decoder. A pre-registered pilot on two cells and seeds
$11$--$20$ scored $25\%$ against $45\%$ for Qwen3-VL-4B, short of the $+4$
episodes the protocol asked for. The difference comes from one cell:
\code{goal\_swap[5]} is $20\%$ for both, and \code{10\_swap[8]} is $30\%$ against
$70\%$. In that cell all four discordant pairs diverge at the first decision,
where Qwen3-VL-4B always names the same moka pot and Sonnet named the other one
in five episodes, none of which succeeded. Tuning the prompt on that cell raised
it from $30\%$ to $60\%$ on development seeds. Confirmation over $40$ episodes
on four cells not used for tuning gave $75\%$ (Qwen3-VL-4B), $80\%$ (Sonnet) and $75\%$
(tuned Sonnet; $p = 0.5$ for the tuned against the untuned arm), so the tuning
did not transfer and the line was stopped. The rule
that helps in \code{10\_swap[8]}, not to change target in mid-episode, hurts in
\code{10\_swap[2]}, where the untuned arm abandons a stuck knob and places the
pot instead. Table~\ref{tab:upper} gives the cost. Opus and GPT models were not
run: keys and budget were not available, and the protocol kept them behind the
pilot's threshold. In our matched PhyAgentOS runs, changing the upper model
from Qwen3-VL-4B to GPT-4o-mini changes success by $1$ percentage point
(Table~\ref{tab:main}).

\begin{table}[ht]
\centering
\caption{\textbf{Latency and token counts in the slow-brain pilot.} Sonnet 5 via Claude Code, $20$ episodes and $107$ calls; Fig.~\ref{fig:teaser}c uses its median.}
\label{tab:upper}
\footnotesize
\setlength{\tabcolsep}{5pt}
\begin{tabular}{@{}lrrr@{}}
\toprule
 & Qwen3-VL-4B & Claude Sonnet 5 & Ratio \\
\midrule
Latency per call, median (ms)   & 1{,}124 & 6{,}683 & 5.9 \\
Latency per call, maximum (ms)  & 1{,}485 & 10{,}348 & 7.0 \\
Prompt tokens per call          & 4{,}036 & 74{,}600 & 18 \\
Prompt tokens per episode       & 21{,}253 & 444{,}102 & 20.9 \\
Slow-brain calls per episode    & 5.35 & 5.35 & 1.0 \\

Calls over the $20$\,s timeout  & 0 & 0 & -- \\
\bottomrule
\end{tabular}
\end{table}

\subsection{Dynamic Physical Harness}
\label{app:retry}

\paragraph{Retry, verification and bounded execution.} In the earlier diagnostic
campaign, PhyAgentOS records three outcomes per episode,
which separate what a retry, a model-based verifier and the underlying policy
each contribute (Table~\ref{tab:phyagentos}; Fig.~\ref{fig:suppl}b). Its retries continue from the
current state with the original instruction, and the step counter restarts on
each attempt, so a retry behaves much like a longer episode. On unperturbed
LIBERO-10, where each attempt was capped at $300$ steps rather than the official
$520$, retries raise the benchmark predicate from $80$ to $92$ of $100$ with
GPT-4o-mini and from $81$ to $96$ with Qwen3-VL-4B, and every recovered episode
finished within $520$ steps in total, so the gain is consistent with additional
budget. On LIBERO-Pro a budget of $50$ verifier calls per cell left most failures
without a retry: $67$ of $320$ and $46$ of $319$ first-attempt failures were
retried, adding $3$ and $2$ successes. The verifier, consulted only after a failed
attempt, accepted $11$ and $22$ of those episodes against the predicate, which
raises recorded success from $20.75\%$ to $23.5\%$ and $26.25\%$ over $400$ episodes; it changes what
is recorded rather than what happens. \sys{} assigns these roles differently. Its
fast brain ends a command on the latched benchmark predicate rather than on a
model's judgment, bounds each command by budget and lease, and retries by
switching capability or recovering within the episode rather than by repeating
the same policy call, as the next subsection shows.

\begin{figure}[t]
\centering
\includegraphics[width=\textwidth]{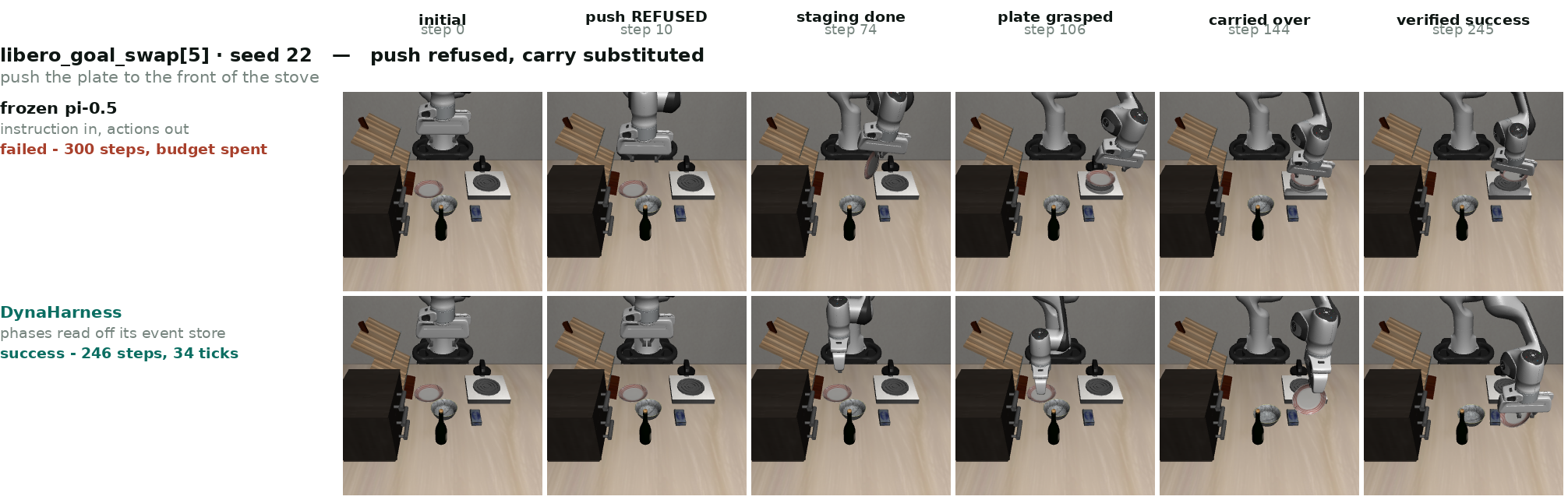}
\caption{\textbf{\code{goal\_swap[5]}, seed 22, both arms}, cut at the same environment steps.}
\label{fig:filmstrip}
\end{figure}

\paragraph{Capability outcomes and composition.}\label{app:qualitative} The retained records contain successful episodes with unsuccessful capability
invocations (Table~\ref{tab:tools}).
\code{turn\_knob\_object} completes only $21\%$ of its invocations, while its source cell improves from $0\%$ to $95\%$ across
development. Failed attempts are retried or replaced within the episode, and \code{keyframe\_recovery} returns the
arm to a recorded pose on all $35$ of its invocations. Fig.~\ref{fig:filmstrip}
shows a substitution. Asked to push a plate, the fast brain refuses
\code{push\_object}, because the jaw is too narrow for the contact the push
requires, and carries the plate instead; the capability records no completion,
yet its cell rises from $0\%$ to $40\%$. Across four scenes recorded through
both arms, the frozen policy fails all four and the agent succeeds in all four.

Composition has limits. \code{slide\_drawer\_object} completed none of its $21$
invocations in the analyzed round, and retrying does not help when a capability
does not complete; the attribution-guided update below raised one drawer cell
from $55\%$ to $100\%$. The verdict itself is the benchmark's own, and our
records contain one case where the two disagree: in \code{10\_swap[9]} the door
capability reported failure while the benchmark scored the episode a success.

\begin{table}[t]
\centering
\caption{\textbf{Evolved capabilities as measured objects.} \emph{Inv.}: dispatches; \emph{Compl.}: completed fraction; \emph{Conversion}: source-cell successes before and after.}
\label{tab:tools}
\scriptsize
\setlength{\tabcolsep}{3pt}
\begin{tabular}{@{}lrrrl@{}}
\toprule
Capability & Inv. & Compl. & Cells & Conversion \\
\midrule
\code{push\_object}          & 1  & 0\,\%   & 1 & \code{goal\_swap[5]} $0\to8$ \\
\code{slide\_drawer\_object} & 21 & 0\,\%   & 2 & \code{10\_task[3]} $0\to0$ \\
\code{turn\_knob\_object}    & 33 & 21\,\%  & 1 & \code{10\_task[2]} $0\to19$ \\
\code{keyframe\_recovery}    & 35 & 100\,\% & 2 & cross-cell \\
\code{swing\_door\_object}   & 1  & 0\,\%   & 1 & \code{10\_swap[9]} $0\to7\to10^{\dagger}$ \\
\bottomrule
\multicolumn{5}{@{}l@{}}{\footnotesize $^{\dagger}$ later rounds, for which only aggregates were retained.}
\end{tabular}
\end{table}

\paragraph{VLA usage.} In the four
successful episodes for which we hold stores, analytic skills account for $85.6$
and $86.3\%$ of the environment steps in the two whose step ledgers cover the
whole episode, and the VLA is not called in any of the four,
while in the four hardest cells the VLA's share ranges from $26.5\%$ to $91.4\%$
and no cell exceeds $35\%$. Across a larger corpus, failures attributed to the
planner layer spend $351$ of their $363$ steps under the VLA, against $107$ of
$321$ for failures attributed to the policy itself: steps fall to the VLA where
the agent has no structure to apply. The converse does not hold, since
\code{10\_swap[3]} keeps most of its steps under analytic skills and still fails
because its drawer capability does not complete. These routing statistics describe the retained sample, which emphasizes hard
cells; they do not estimate the effect of removing the VLA.

\subsection{Attribution-Driven Self-Evolution}

\begin{figure}[t]
\centering
\includegraphics[width=\textwidth]{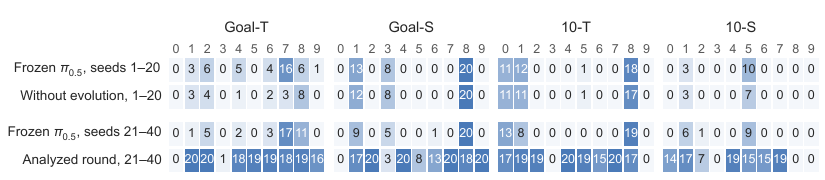}
\caption{\textbf{Successes of 20 per cell, four arms}; the early seeds $1$--$20$ block
(top) and the development block (bottom) are compared only within a block.}
\label{fig:percell}
\end{figure}

\paragraph{Development changes across cells.} The evolved agent differs from the harness without evolution by what was
added over the evolution rounds: five new capabilities and changes to existing layers,
including the latched verdict, receptacle slots, a handle grasp and a transport
corridor. Its gains are local rather than uniform (Fig.~\ref{fig:percell}). In
the analyzed round, against the frozen policy on the same seeds, $27$ of $40$
cells gain five or more successes, $3$ lose a few, and $7$ remain at zero. Four of
the zero cells and two nearly-zero cells involve a drawer, and the four
drawer cells in the Goal suites account for $76$ of the $111$ episodes those
suites still lose. What the reported agent still loses is local too: of the $207$
episodes lost in its second measurement ($74.1\%$), $97$ involve a drawer, $43$
a stove and $30$ a microwave, consistent with changes that each target the layer
and the physical situation that failed.

\paragraph{Updating a capability following attribution.}\label{app:drawer}
Attribution assigned
the drawer cells to the capability layer. The implicated capability was then
probed and revised. A probe of
\code{goal\_swap[0]}, which asks for the middle drawer to be opened, found three
physical causes: the forearm wedged against the wine rack while pulling, the
gripper's closure limit rejected about a third of real grasps on the $15.4$\,mm
handle, and after contact the hand often sat too high or too far to close on the
handle. Two single-factor changes followed, each a default-off setting stated in
these quantities (Table~\ref{tab:drawer}). The first shifts the grasp $20$\,mm
along the handle and widens the closure limit; on seeds $40$--$59$ it raised the cell from $30\%$ to $85\%$, with $11$ pairs won and
none lost. In its initial paired Goal confirmation, both arms scored $76.0\%$:
the local mechanism criterion passed, but the predeclared total-score improvement
criterion did not. The candidate was retained for further development, not
counted as passing the full admission gate; the thresholds were not revised.
The second change re-seats the hand after a
misaligned contact and waits for two consecutive stalled pulls before giving up.
A confirmation on all Goal-T and Goal-S episodes of the development block, under a
reachability rule declared before any episode was run, raised the cell from
$55\%$ to $100\%$ and Goal success from $75.25\%$ to $78.0\%$ over $400$ episodes,
with no
discordant pair among the other $17$ unreachable cells. Table~\ref{tab:drawer}
reports this later confirmation, not the first change's initial $76.0\%$ test.
The reported agent includes the combined configuration. Local mechanism evidence
supports continued development; formal admission additionally requires the paired
and broader regression checks in Section~\ref{sec:evolution}.
On the unperturbed suites they are
inert: no cell of the four suites reaches them, and a paired comparison of the two
builds over the same $2{,}000$ episodes shows no difference ($p = 0.49$); the
absolute scores come from a tree in which the orchestration tick limit stopped
about $1{,}070$ episodes per arm, so they are not quoted.

\begin{table}[t]
\centering
\caption{\textbf{The two drawer changes}, $400$ episodes per arm on seeds $21$--$40$, paired by task and seed; unreachable cells are not scored.}
\label{tab:drawer}
\footnotesize
\setlength{\tabcolsep}{5pt}
\begin{tabular}{@{}lrrrl@{}}
\toprule
Cell & Agent & First change & Both & Paired outcome \\
\midrule
\code{goal\_swap[0]} (reachable)  & 11 & 18 & \textbf{20} & 2 pairs won, none lost \\
\code{goal\_swap[3]} (unreachable)& 2  & 3  & 3  & 2 against 2 \\
\code{goal\_task[7]} (unreachable)& 18 & 17 & 19 & within repeat noise \\
17 further cells (unreachable)    & 270 & 270 & 270 & no discordant pair \\
\midrule
Goal-T and Goal-S ($/400$)        & 301 & 308 & \textbf{312} & \\
\bottomrule
\end{tabular}
\end{table}

\paragraph{A rejected candidate.} A per-cell comparison identified a regression
after a local candidate passed its gate. One admitted candidate generalized
a cavity entry built for a microwave to every region resting on a fixture body,
including a stove's flat hob whose cavity was $5$\,mm tall. In the next full
evaluation, five cells lost $79$ successes between them while every other shared
cell moved by at most one (Table~\ref{tab:negative}). All five collapsed cells
place an object onto the hob, whereas stove cells that only turn the knob or push
along the surface were unaffected, so the change had turned a placement into an
insertion. Episodes left to the policy rose from $166$ to $226$, the signature of
a removed analytic route (Fig.~\ref{fig:composition}). The candidate was reverted
rather than patched, and the following evaluation restored all five cells.

\begin{table}[t]
\centering
\caption{\textbf{The rejected candidate over 39 shared cells}, successes of $20$.}
\label{tab:negative}
\scriptsize
\setlength{\tabcolsep}{3pt}
\begin{tabular}{@{}llccr@{}}
\toprule
Cell & Instruction & Prom. & Rej. & $\Delta$ \\
\midrule
\code{goal\_task[1]} & plate on stove   & 20 & 0 & $-20$ \\
\code{10\_task[2]}   & stove on + pan   & 19 & 1 & $-18$ \\
\code{goal\_swap[1]} & bowl on stove    & 17 & 0 & $-17$ \\
\code{10\_task[8]}   & moka pot on stove& 17 & 0 & $-17$ \\
\code{10\_swap[2]}   & stove on + moka  & 7  & 0 & $-7$  \\
\midrule
\code{10\_swap[9]}   & mug in microwave & 0  & 5 & $+5$  \\
\multicolumn{2}{@{}l}{33 further shared cells} & \multicolumn{3}{c}{within $\pm1$} \\
\midrule
\multicolumn{2}{@{}l}{Total over 39 shared cells} & 540 & 466 & $-74$ \\
\bottomrule
\end{tabular}
\end{table}

\paragraph{A local improvement is not a global one.} Validating a change on the cells it targets is not enough. The rejected round
above also contained a genuine local gain, a hinged-door capability that raised
the microwave cell from $0\%$ to $25\%$; judged by the round total, the two
changes would have been kept or discarded together, while the per-cell comparison
kept one and reverted the other. A second candidate, which widened the conditions
for a mirror recovery, won its paired comparison on six targeted cells, $61.7\%$
against $55.8\%$, with no net change on four regression cells, yet the following
full evaluation scored $70.75\%$ against $72.75\%$ over $800$ episodes, with losses in cells it
did not target, and the incumbent was kept (Fig.~\ref{fig:suppl}c;
Appendix~\ref{app:twostage}). A
matched comparison on the affected suite scored $76.5\%$ for both builds.
The earlier Goal-S loss therefore did not reproduce. This candidate documents
a decision not to promote after screening, rather than a confirmed regression.

\section{Case Studies}
\label{app:cases}

This section illustrates refusal, substitution and capability execution in
selected \sys{} episodes. Frames are aligned by environment step where a
frozen-policy control is shown. The cases complement the aggregate results;
source-cell counts and invocation statistics are in Table~\ref{tab:tools}.
Diagnostic baseline runs are described in Appendix~\ref{sec:ranbaselines}.


\begin{figure}[!htb]
\centering
\begin{ourbox}{DynaHarness: an infeasible push is refused and replaced within the episode}
\setlength{\cfw}{0.162\linewidth}
\noindent
\cframe{initial, step 0}{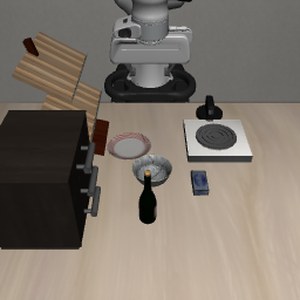}\hfill
\cframe[okgreen]{push refused, 10}{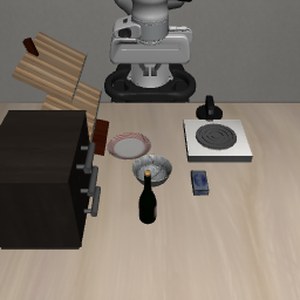}\hfill
\cframe{staging done, 74}{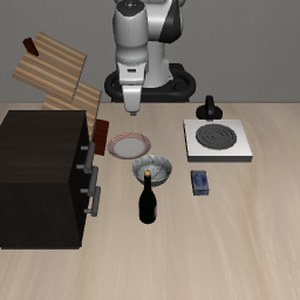}\hfill
\cframe{plate grasped, 106}{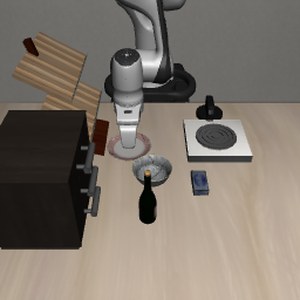}\hfill
\cframe{carried over, 144}{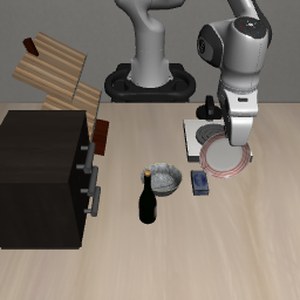}\hfill
\cframe[okgreen]{completed, 245}{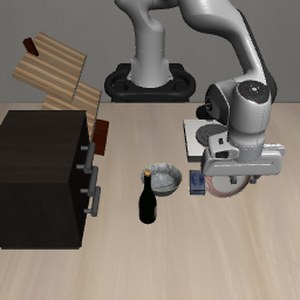}\\[4pt]
\begin{minipage}[t]{0.53\linewidth}\small
\textbf{Episode:} \code{goal\_swap[5]}, seed $22$, \emph{push the plate to the
front of the stove}; budget $300$ steps.\\[2pt]
\textbf{Slow brain}: names \code{push\_object} for the plate.\\[1pt]
\textbf{Fast brain}, step $10$: \good{refuses the push, because the jaw is too
narrow for the contact the push requires}, and records the reason for the
refusal.\\[1pt]
\textbf{Fast brain}: substitutes an eligible pick and place; analytic stages
stage, grasp, carry and place the plate.\\[1pt]
\textbf{Verdict}, step $245$: the latched benchmark predicate becomes true and
ends the command.\\[1pt]
\textbf{Frozen \pizero{}}, same seed: \fail{fails after spending
all $300$ steps of the budget}.
\end{minipage}\hfill
\begin{minipage}[t]{0.44\linewidth}\small
\textbf{Diagnosis:}\\[2pt]
\good{The episode succeeds in $246$ steps, with $34$ decisions by the fast
brain.}\\[2pt]
\textbf{Mechanism.} Refusal is a returned value with a reason
(Eq.~\ref{eq:ground}), so a capability the scene does not support is replaced
inside the episode instead of being executed until the budget runs out. The
command ends on the latched predicate (Eq.~\ref{eq:verify}), not on a model's
judgment.\\[2pt]
\textbf{Effect.} \code{push\_object} records no completion, yet its source cell
rises from $0\%$ to $40\%$ on the same seeds (Table~\ref{tab:tools}).
\end{minipage}
\end{ourbox}
\caption{\textbf{Refusal and substitution} of an infeasible capability by the fast brain.}
\label{fig:case_ours_push}
\end{figure}

\begin{figure}[!htb]
\centering
\begin{ourbox}{DynaHarness: an admitted door capability completes a long-horizon task}
\setlength{\cfw}{0.162\linewidth}
\noindent
\cframe{initial, step 0}{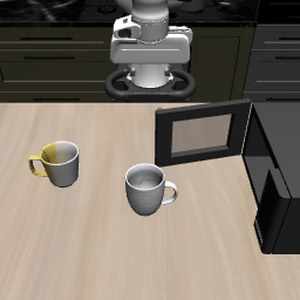}\hfill
\cframe{staging done, 82}{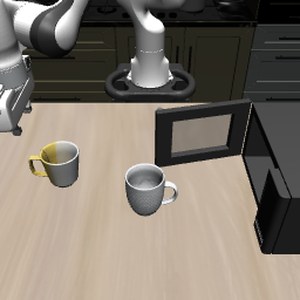}\hfill
\cframe{jaw closed, 130}{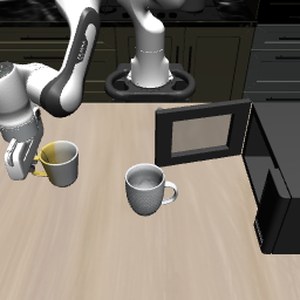}\hfill
\cframe{at the cavity, 208}{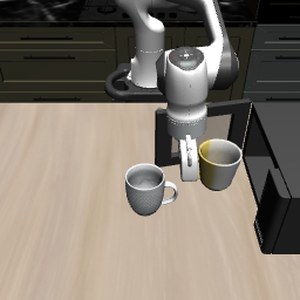}\hfill
\cframe{released inside, 250}{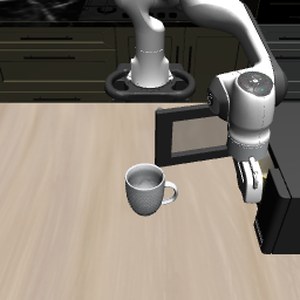}\hfill
\cframe[okgreen]{door shut, 519}{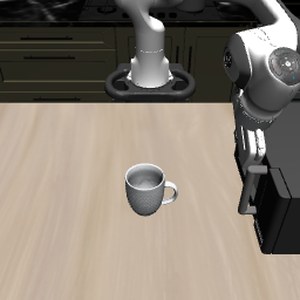}\\[3pt]
\cframe{\emph{frozen \pizero{}}, 0}{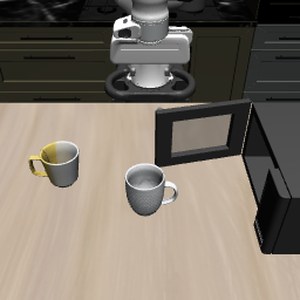}\hfill
\cframe{82}{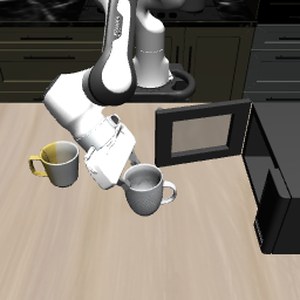}\hfill
\cframe{130}{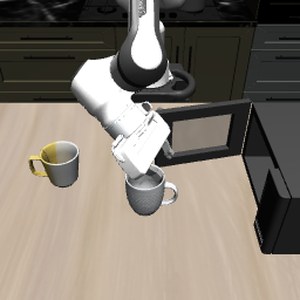}\hfill
\cframe{208}{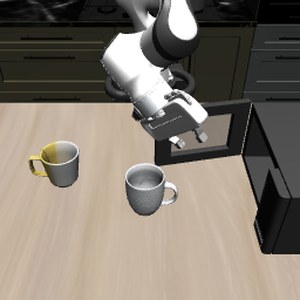}\hfill
\cframe{250}{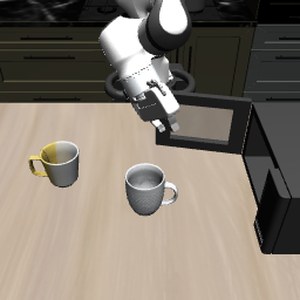}\hfill
\cframe[failred]{519, budget spent}{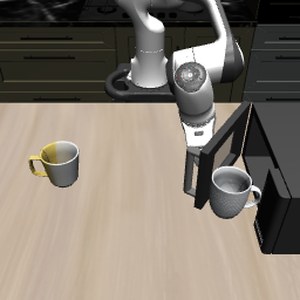}\\[4pt]
\begin{minipage}[t]{0.53\linewidth}\small
\textbf{Episode:} \code{10\_swap[9]}, seed $21$, \emph{put the yellow and white
mug in the microwave and close it}; the episode budget is $520$ steps.\\[1pt]
\textbf{Upper row, \sys{}}: analytic stages stage the arm and grasp the mug,
carry it into the cavity and release it; the hinged-door capability
\code{swing\_door} then closes the door, and the latched predicate becomes true
at step $519$.\\[1pt]
\textbf{Lower row, frozen \pizero{}}, cut at the same steps: \fail{the task is
not completed within the $520$-step budget}.
\end{minipage}\hfill
\begin{minipage}[t]{0.44\linewidth}\small
\textbf{Diagnosis:}\\[2pt]
\good{The episode succeeds in $520$ steps, with $54$ decisions by the fast
brain.}\\[2pt]
\textbf{Mechanism.} Before the change, attribution found that the plan step
\emph{close microwave} compiled to no command, so the policy owned all $520$
steps. The candidate \code{swing\_door}, which declares the hinge and the
cavity behind it, passed the paired gate of Eq.~\ref{eq:gate}.\\[2pt]
\textbf{Effect.} The cell moved from $0\%$ to $35\%$ on the same seeds and
later to $50\%$ (Table~\ref{tab:tools}).
\end{minipage}
\end{ourbox}
\caption{\textbf{An evolved capability at work}, beside the frozen policy on the same seed.}
\label{fig:case_ours_door}
\end{figure}

\begin{figure}[!htb]
\centering
\begin{ourbox}{DynaHarness: long-horizon tasks finish inside the budget}
\setlength{\cfw}{0.162\linewidth}
\noindent
\cframe{initial, step 0}{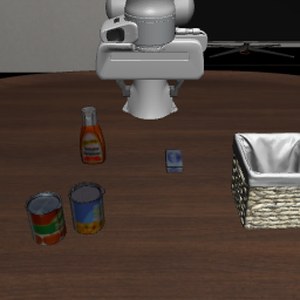}\hfill
\cframe{staging done, 40}{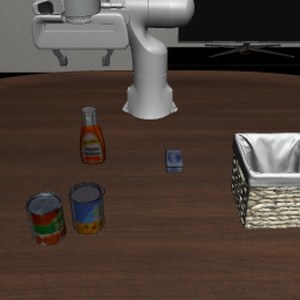}\hfill
\cframe{1st jaw closed, 84}{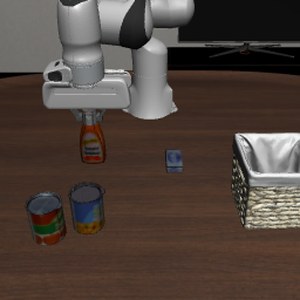}\hfill
\cframe{1st in its slot, 222}{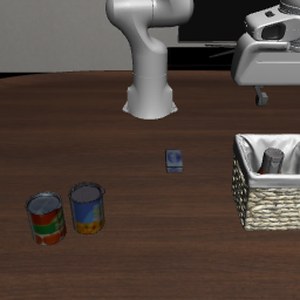}\hfill
\cframe{2nd jaw closed, 308}{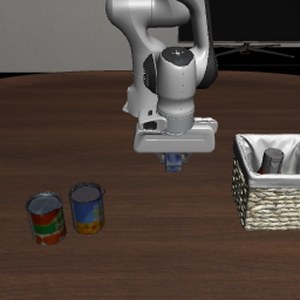}\hfill
\cframe[okgreen]{both placed, 409}{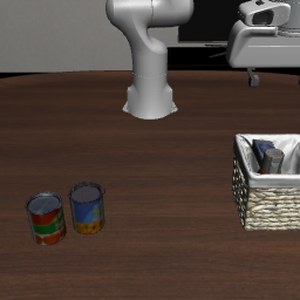}\\[3pt]
\cframe{initial, step 0}{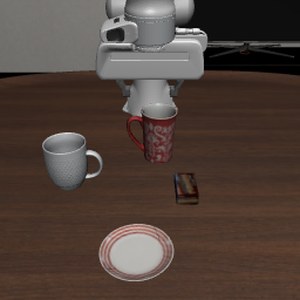}\hfill
\cframe{staging done, 70}{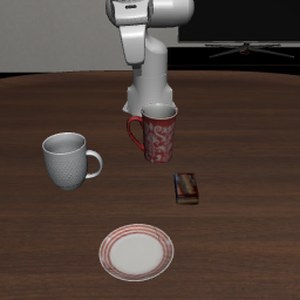}\hfill
\cframe{mug grasped, 114}{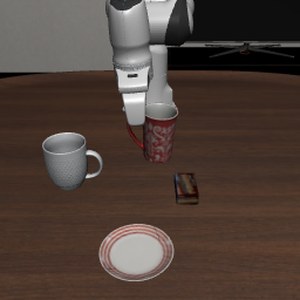}\hfill
\cframe{mug on plate, 197}{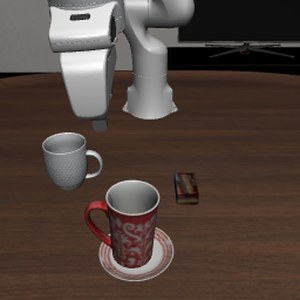}\hfill
\cframe{pudding grasped, 325}{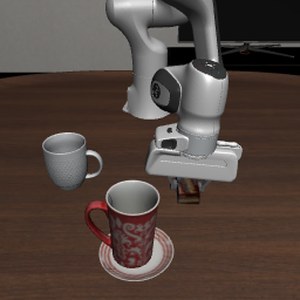}\hfill
\cframe[okgreen]{placed clear, 436}{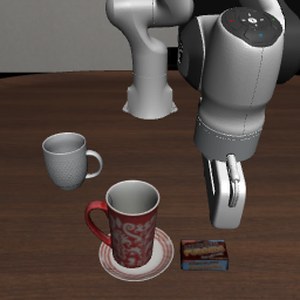}\\[4pt]
\begin{minipage}[t]{0.53\linewidth}\small
\textbf{Upper row:} \code{10\_task[7]}, seed $21$, \emph{put both the ketchup
and the cream cheese box in the basket}; budget $520$ steps. Receptacle slots
give each object its own place in the basket, and the verdict latches once both
objects have been placed.\\[2pt]
\textbf{Lower row:} \code{10\_task[6]}, seed $23$, \emph{put the red mug on the
plate and the chocolate pudding right of it}; budget $520$ steps. Three motion
rules (a climb before transport, a corridor over the table and a placement
clearance) move the two objects in sequence.\\[2pt]
\textbf{Frozen \pizero{}}, same seeds: \fail{both episodes fail after spending
the whole budget}.
\end{minipage}\hfill
\begin{minipage}[t]{0.44\linewidth}\small
\textbf{Diagnosis:}\\[2pt]
\good{Success in $410$ and $437$ of $520$ steps, with $55$ and $52$ fast-brain
decisions.}\\[2pt]
\textbf{Mechanism.} Analytic stages spend the steps that measured geometry
determines, so a two-object task leaves budget to spare instead of running out
while a stage fails to converge (Section~\ref{sec:episodes}). In these successes analytic
skills account for $85.6\%$ and $86.3\%$ of the steps and the VLA is not
called (Fig.~\ref{fig:mech}f).
\end{minipage}
\end{ourbox}
\caption{\textbf{Two long-horizon episodes} whose frozen-policy counterparts fail on the same seeds.}
\label{fig:case_ours_long}
\end{figure}

\section{Per-Task Results}
\label{app:pertask}

Table~\ref{tab:pertask} breaks the four LIBERO-Pro cells down by task. The
frozen-policy references disagree task by task and not only on average. Against
our early seeds $1$--$20$ block, the reference in Zetta's evaluation differs by $50$ points or
more on nine of the $40$ tasks, in both directions: \code{goal\_task[1]} succeeds
on $95\%$ of seeds there and $15\%$ here, and \code{10\_task[0]} on $5\%$ there
and $55\%$ here. At round 39, before the drawer changes, the agent succeeds on at
least $80\%$ of seeds in $27$ of
$40$ tasks and on none in five, four of which involve a drawer while the fifth
closes a microwave.

\begin{table}[t]
\centering
\caption{\textbf{Per-task success (\%) on LIBERO-Pro.} $\dagger$Seeds $1$--$20$, Table~3 of \citet{zetta2026}; $\star$ours, \sys{} at round 39.}
\label{tab:pertask}
\scriptsize
\setlength{\tabcolsep}{3.4pt}
\begin{tabular}{@{}llccccccccccc@{}}
\toprule
Cell & Method & 0 & 1 & 2 & 3 & 4 & 5 & 6 & 7 & 8 & 9 & Avg. \\
\midrule
\multirow{5}{*}{Goal-T}
 & \pizero{}~$\dagger$~\citep{zetta2026}   & 0 & 95 & 10 & 0 & 100 & 0 & 20 & 80 & 5 & 0 & 31.0 \\
 & Zetta~$\dagger$~\citep{zetta2026}       & 80 & 100 & 95 & 80 & 100 & 100 & 95 & 95 & 80 & 100 & 92.5 \\
 & \pizero{}~$\star$ seeds 1--20          & 0 & 15 & 30 & 0 & 25 & 0 & 20 & 80 & 30 & 5 & 20.5 \\
 & \pizero{}~$\star$ seeds 21--40         & 0 & 5 & 25 & 0 & 10 & 0 & 15 & 85 & 55 & 0 & 19.5 \\
 & \sys{}~$\star$ 21--40                  & 0 & 100 & 100 & 0 & 90 & 95 & 95 & 90 & 95 & 80 & 74.5 \\
\midrule
\multirow{5}{*}{Goal-S}
 & \pizero{}~$\dagger$~\citep{zetta2026}   & 0 & 60 & 0 & 45 & 0 & 0 & 0 & 100 & 100 & 75 & 38.0 \\
 & Zetta~$\dagger$~\citep{zetta2026}       & 90 & 65 & 80 & 85 & 95 & 95 & 100 & 100 & 100 & 80 & 89.0 \\
 & \pizero{}~$\star$ seeds 1--20          & 0 & 65 & 0 & 40 & 0 & 0 & 0 & 0 & 100 & 0 & 20.5 \\
 & \pizero{}~$\star$ seeds 21--40         & 0 & 45 & 0 & 25 & 0 & 0 & 5 & 0 & 100 & 0 & 17.5 \\
 & \sys{}~$\star$ 21--40                  & 55 & 80 & 100 & 10 & 100 & 50 & 70 & 100 & 95 & 100 & 76.0 \\
\midrule
\multirow{5}{*}{10-T}
 & \pizero{}~$\dagger$~\citep{zetta2026}   & 5 & 95 & 95 & 0 & 0 & 80 & 85 & 75 & 65 & 0 & 50.0 \\
 & Zetta~$\dagger$~\citep{zetta2026}       & 35 & 95 & 100 & 0 & 25 & 100 & 95 & 80 & 100 & 0 & 63.0 \\
 & \pizero{}~$\star$ seeds 1--20          & 55 & 60 & 0 & 0 & 0 & 5 & 0 & 0 & 90 & 0 & 21.0 \\
 & \pizero{}~$\star$ seeds 21--40         & 65 & 40 & 0 & 0 & 0 & 0 & 0 & 0 & 95 & 0 & 20.0 \\
 & \sys{}~$\star$ 21--40                  & 85 & 95 & 95 & 0 & 100 & 95 & 90 & 100 & 100 & 0 & 76.0 \\
\midrule
\multirow{5}{*}{10-S}
 & \pizero{}~$\dagger$~\citep{zetta2026}   & 0 & 35 & 0 & 0 & 5 & 50 & 0 & 0 & 0 & 0 & 9.0 \\
 & Zetta~$\dagger$~\citep{zetta2026}       & 90 & 50 & 95 & 75 & 15 & 65 & 5 & 0 & 0 & 5 & 40.0 \\
 & \pizero{}~$\star$ seeds 1--20          & 0 & 15 & 0 & 0 & 0 & 50 & 0 & 0 & 0 & 0 & 6.5 \\
 & \pizero{}~$\star$ seeds 21--40         & 0 & 30 & 5 & 0 & 0 & 45 & 0 & 0 & 0 & 0 & 8.0 \\
 & \sys{}~$\star$ 21--40                  & 90 & 90 & 35 & 0 & 95 & 65 & 90 & 95 & 40 & 45 & 64.5 \\
\bottomrule
\end{tabular}
\end{table}

\section{Baseline Records}
\label{sec:ranbaselines}

\paragraph{PhyAgentOS.} We ran PhyAgentOS over the \pizero{} LIBERO checkpoint
configuration in an earlier diagnostic campaign with GPT-4o-mini and with
Qwen3-VL-4B, the same local 4-bit build that \sys{} uses as its slow brain. In these runs the upper model submits each cell's episodes once and
is not asked to plan per episode. Within an episode, the same model acts as a
verifier after an attempt whose benchmark predicate is false: it sees the
agent-view and wrist images at the start and end of the attempt together with the
predicate's failure, and it answers \emph{success}, which ends the episode as a
success, \emph{replan}, which starts another attempt from the current state with
the original instruction, or \emph{failure}. Each episode has at most three
attempts, and each cell at most $50$ verifier calls. On LIBERO every attempt is
capped at $300$ steps, including LIBERO-10, whose official budget is $520$; on
LIBERO-Pro, seed $s$ sets both the initial state and the environment seed, under
the official budgets. Table~\ref{tab:phyagentos} gives the three outcomes per
suite or cell and Table~\ref{tab:phycost} the execution statistics. Six of the
seven schema-invalid Qwen responses were \emph{success} verdicts rejected for a
missing field. Each diagnostic condition was run once.

\begin{table}[t]
\centering
\caption{\textbf{PhyAgentOS run by us}, successes of $100$ per suite or cell.}
\label{tab:phyagentos}
\footnotesize
\setlength{\tabcolsep}{3.5pt}
\begin{tabular}{@{}lrrrrrr@{}}
\toprule
 & \multicolumn{3}{c}{GPT-4o-mini} & \multicolumn{3}{c}{Qwen3-VL-4B} \\
\cmidrule(lr){2-4}\cmidrule(l){5-7}
Suite or cell & F & Fi & V & F & Fi & V \\
\midrule
Spatial                   & 100 & 100 & 100 & 98 & 98 & 99 \\
Object                    & 99  & 100 & 100 & 99 & 99 & 99 \\
Goal                      & 98  & 98  & 98  & 98 & 98 & 99 \\
Long                      & 80  & 92  & 95  & 81 & 96 & 97 \\
LIBERO ($/400$)           & 377 & 390 & 393 & 376 & 391 & 394 \\
\midrule
Goal-T                    & 19 & 19 & 23 & 21 & 22 & 34 \\
Goal-S                    & 36 & 36 & 39 & 34 & 34 & 41 \\
10-T                      & 17 & 17 & 20 & 18 & 18 & 20 \\
10-S                      & 8  & 11 & 12 & 8  & 9  & 10 \\
LIBERO-Pro ($/400$)       & 80 & 83 & 94 & 81 & 83 & 105$^{\dagger}$ \\
\bottomrule
\end{tabular}
\par\vspace{2pt}{\footnotesize F, Fi: predicate after the first and the last of three
attempts; V: success the system recorded. $^{\dagger}$Plus $7$ schema-invalid
verifier replies.}
\end{table}

\begin{table}[t]
\centering
\caption{\textbf{Execution statistics of our PhyAgentOS runs}, from their episode records.}
\label{tab:phycost}
\footnotesize
\setlength{\tabcolsep}{4pt}
\begin{tabular}{@{}lrrrr@{}}
\toprule
 & \multicolumn{2}{c}{LIBERO} & \multicolumn{2}{c}{LIBERO-Pro} \\
\cmidrule(lr){2-3}\cmidrule(l){4-5}
 & GPT & Qwen & GPT & Qwen \\
\midrule
Attempts per episode, mean           & 1.07 & 1.04 & 1.30 & 1.16 \\
Episodes retried                     & 21   & 16   & 67   & 46   \\
Verifier calls                       & 39   & 25   & 200  & 200  \\
Environment steps per episode, mean  & 168  & 158  & 501  & 442  \\
Policy calls per episode, mean       & 34.0 & 32.0 & 100.2 & 88.4 \\
Verifier latency per call (s), mean  & 5.4  & 4.2  & 4.9  & 4.3  \\
Wall clock (h)                       & 0.83 & 0.77 & 2.40 & 2.19 \\
\bottomrule
\end{tabular}
\end{table}

\paragraph{Zetta.} Table~\ref{tab:zetta4b} gives our Qwen3-VL-4B runs stage by
stage and the GPT-5.6-sol candidate. These diagnostic runs examine recovery and candidate admission in the released
code. The GPT-5.6-sol candidate is evaluated against the bare policy over $20$
episodes of $300$ policy steps each.
With Qwen3-VL-4B, the online recovery agent has to read a current camera image
before deciding; across three invocations it
wrote the image request as text instead of a tool call and failed closed. The
offline diagnosis attached three visual-evidence claims to images that were not
delivered and was rejected by its validator. In an episode where the agent did
act, it accepted a recovery proposal with confidence $0.95$, and the recovery
reported completion after $78$ steps while the drawer's joint target was not met.
At the evolution stage, $8$ of $9$ candidate proposals failed validation, for not
citing the rejected gate's evidence or for repeating the rejected mechanism, and
the three same-seed gates all failed, since the candidate left the trajectory
unchanged.

\begin{table}[t]
\centering
\caption{\textbf{Zetta's released code on drawer tasks}: where each Qwen3-VL-4B stage stopped (top) and a GPT-5.6-sol candidate against the bare policy (bottom).}
\label{tab:zetta4b}
\scriptsize
\setlength{\tabcolsep}{3pt}
\begin{tabular}{@{}p{0.27\columnwidth}p{0.67\columnwidth}@{}}
\toprule
Stage & Recorded outcome \\
\midrule
Online recovery agent & 3 invocations; image read requested as text, 0 tool
calls; failed closed with no environment write \\
Offline diagnosis & 3 visual-evidence claims citing images that were not delivered;
diagnosis rejected by the validator \\
Recovery execution & proposal accepted at confidence 0.95; reported
\emph{completed} after 78 steps; joint target unmet; episode failed \\
Candidate proposal & 8 of 9 proposals failed validation \\
Same-seed gate & 3 gates, 0 successes in either arm; none passed \\
\midrule
GPT-5.6-sol candidate & 13/20 against the bare policy's 6/20; discordant pairs 8 and 1,
exact sign test $p = 0.039$ \\
Tool use & \code{privileged\_pick\_place} in 14 of 20 candidate episodes and 10
of its 13 successes; none in the bare policy \\
\bottomrule
\end{tabular}
\end{table}

\paragraph{ENPIRE and ASPIRE with Qwen3-VL-4B.} We ran ENPIRE's released
agent loop with Qwen3-VL-4B writing one Python policy per LIBERO-Pro task
(seed $0$, $310$ steps, \pizero{} called through \code{pi05.predict}). The run
summary records $3$ successes in $40$ tasks (Goal-T $1$, Goal-S $2$, LIBERO-10
$0$) and attributes $28$ failures to interface errors in the generated code, $7$
to the step limit, and one each to a request timeout and an empty initial state.
Raw files survive for $18$ of the $40$ tasks, all consistent with the summary
and none of them a success, so the total rests on the summary. Of the $18$
recovered policies, $14$ call names defined nowhere in the policy or its inputs
and $4$ only pass the observation to \pizero{}; a separate session repeated one
observation call $20$ times without acting. We ran
ASPIRE through Claude Code with Qwen3-VL-4B-Instruct on six LIBERO and
LIBERO-Pro suites. No run executed or scored an episode: replies that described
commands without calling a tool ended the loop, and one subagent searched $187$
times for a directory that does not exist before the coordinator reported a
background run that did not exist. These first runs document adaptation failures. A second campaign, summarized in
\code{aspire\_phyagent.xlsx} and \code{ENPIRE\_LIBERO\_Pro\_success\_rates.xlsx},
covers the four LIBERO-Pro cells with $200$ episodes each for PhyAgentOS
(GPT-4o-mini $21.3\%$, Qwen3-VL-4B $20.3\%$), ENPIRE ($5.4\%$; Goal-T $10.0\%$,
Goal-S $11.5\%$, LIBERO-10 $0\%$), Zetta with Qwen3-VL-4B (no success) and
ASPIRE with Qwen3-VL-4B, whose runs again ended without an executed episode.

\paragraph{Harness VLA agent.} Table~\ref{tab:hvla} gives three diagnostic
episodes of the Harness VLA agent on the LIBERO-Pro Spatial-T cell. A Codex
agent drives segmentation, back-projection, motion and grasp tools turn by turn,
including \code{pi0\_pick} and \code{pi0\_doubled}. Before acting, it reads per-task notes that
include the task text, known failure modes and a seed-0 run of the same perturbed
task, and after each action it observes the benchmark's termination signal. The
successful episode recovered from an off-plate release by re-measuring the plate
center from an unoccluded mask and grasping again. One failure left a bowl upright
on the plate rim after several pushing corrections. In the other, by the agent's
own account, it lost the identity of two similar bowls once they occluded each
other, and it used the termination signal to reason about which bowl was the
target. Each turn that moves the arm issues one arm command, and each turn resends
a context of about $10^5$ input tokens, $94$ to $95\%$ of which was cached.

\begin{table}[t]
\centering
\caption{\textbf{Harness VLA agent, three Spatial-T episodes.} Tokens in millions.}
\label{tab:hvla}
\footnotesize
\setlength{\tabcolsep}{3.2pt}
\begin{tabular}{@{}llrrrrr@{}}
\toprule
Task, seed & Outcome & Turns & Tools & Policy & Tokens & Time (s) \\
\midrule
1, 1 & success & 31 & 62 & 3 & 3.32 & 348 \\
1, 3 & failure & 37 & 70 & 4 & 4.51 & 505 \\
0, 5 & failure & 33 & 81 & 4 & 3.27 & 448 \\
\bottomrule
\end{tabular}
\end{table}

\paragraph{Harness VLA with the same model and the same policy.} We then ran the
released agent over our four cells with Qwen3-VL-4B in place of the coding agent
and the frozen public \pizero{} as the low-level policy, $200$ episodes per cell.
It completed all $800$ episodes and succeeded in $48$: $28$ on Goal-T, $14$ on
Goal-S, $6$ on 10-T and none on 10-S. This is the matched Qwen3-VL-4B plus
frozen-\pizero{} comparison in Table~\ref{tab:main}, where \sys{} reaches
$74.25\%$ against Harness VLA's $6.0\%$.

\paragraph{CaP-Agent0 with Qwen3-VL-4B.}
Our CaP-X adaptation used Qwen3-VL-4B-Instruct with the pointing model disabled
and scored $5.0\%$: one success each in Goal-T and Goal-S. Two retained episode
records show generated-code failures, and a follow-up prompt in one contains an
unfilled task template. These runs characterize this adaptation and are excluded
from Table~\ref{tab:main}.

\section{Two-Stage Admission of a Recovery Candidate}
\label{app:twostage}

Candidate C2 widened the conditions under which a mirror recovery fires. On a
$200$-episode paired benchmark of six targeted cells and four regression cells at
$20$ seeds each, the incumbent scored $55.8\%$ on the targeted cells, an
intermediate candidate C1 $58.3\%$ and C2 $61.7\%$, and the regression cells
showed no net change for either candidate (Table~\ref{tab:twostage}). Under C2
the mirror fired $36$ times against $12$ under C1 and reached all six targeted
cells rather than three; in $23$ firings the gripper kept the object, and $7$ of
those episodes succeeded. Two cells did not move: \code{goal\_swap[5]} fired it
$10$ times and stayed at $50\%$, and \code{10\_swap[3]} fired it $6$ times and
stayed at $0\%$, so for both the grasp was no longer the stage that failed. The
pre-declared confirmation then ran C2 as a full $800$-episode round. The incumbent
had scored $72.75\%$ and C2 scored $70.75\%$, with $13$ cells moving, $6$ episodes
gained and $22$ lost, $20$ of the losses inside \code{goal\_swap} and the largest
in \code{goal\_swap[4]} ($20 \to 14$) and \code{goal\_swap[7]} ($20 \to 15$),
neither of them targeted. The incumbent was kept. A later matched comparison of
the two builds on the same $200$ Goal-S task and seed pairs found $76.5\%$ for
each, with $2$ pairs won only by each side, so the Goal-S loss did not
reproduce.

\begin{table}[t]
\centering
\caption{\textbf{Two-stage admission of candidate C2}: paired mechanism benchmark, full round and matched Goal-S comparison, top to bottom.}
\label{tab:twostage}
\footnotesize
\setlength{\tabcolsep}{5pt}
\begin{tabular}{@{}lrrr@{}}
\toprule
Paired mechanism benchmark & Incumbent & C1 & C2 \\
\midrule
\code{goal\_task[9]}      & 16 & 16 & 18 \\
\code{10\_task[0]}        & 17 & 17 & 19 \\
\code{goal\_swap[1]}      & 16 & 18 & 18 \\
\code{10\_swap[8]}        & 8  & 9  & 9  \\
\code{goal\_swap[5]}      & 10 & 10 & 10 \\
\code{10\_swap[3]}        & 0  & 0  & 0  \\
Targeted total ($/120$)   & 67 & 70 & 74 \\
Regression cells, net     & n/a & 0 & 0 \\
\midrule
Full round ($/200$ per cell) & Incumbent & & C2 \\
\midrule
Goal-T                    & 149 & & 151 \\
Goal-S                    & 152 & & 133 \\
10-T                      & 152 & & 153 \\
10-S                      & 129 & & 129 \\
Total ($/800$)            & 582 & & 566 \\
\midrule
Matched Goal-S ($/200$)   & 153 & & 153 \\
\bottomrule
\end{tabular}
\end{table}

\section{Limitations}
\label{sec:limitations}

\paragraph{Evaluation scope.}
Post-selection tests cover new initial states of the same $40$ LIBERO-Pro
cells and $261$ untouched official states across $31$ cells. They test initial-state
transfer; new task families and physical domains require separate evaluations.
Standard-LIBERO and LIBERO-Plus configuration records are in
Appendix~\ref{app:eval}; LIBERO-Plus has one trial per task without a
frozen-policy control. The executor comparisons use development states and
evaluate execution mechanisms jointly. A2static retains the one-step planner
interface; A2seq also changes planning horizon and output schema. The A2seq single-host
run and historical multi-host control also differ in date and machine;
unseeded planner and policy stochasticity remain nuisance factors
(Appendix~\ref{app:seq}).

\paragraph{Improvement and deployment.}
Failure attribution and paired admission automate the diagnostic and evaluation
steps surrounding fault probing and targeted revision.
The deployed fast brain is deterministic; the learned variant matches its
success count in the reported comparison (Appendix~\ref{app:learned}).
LIBERO completion uses the benchmark predicate, while physical deployment
requires a separately specified completion interface and scene description.

\paragraph{Records and coverage.}
The earlier step-share case analyses cover $8$ of $40$ cells. New mechanism
and post-selection event-store coverage is reported in
Table~\ref{tab:routing_new}. The final development score is an archived
aggregate; a separate remeasurement supplies the historical paired comparison. Some earlier
rounds and baseline campaigns retain only summaries. The no-evolution arm has
no attribution field, and the drawer confirmation covers the two Goal cells.
Appendices~\ref{app:eval}, \ref{app:results} and~\ref{sec:ranbaselines} describe
run-specific processing and record availability.

\end{document}